\documentclass[journal]{IEEEtran}
\usepackage{lineno,hyperref}
\usepackage{subfigure}
\usepackage{epsfig}

\usepackage{amssymb}
\usepackage{graphicx}
\usepackage{algorithm}
\usepackage{amsmath}
\usepackage{multirow}
\usepackage{diagbox}
\usepackage{colortbl}
\usepackage{booktabs}
\newcommand{\RNum}[1]{\uppercase\expandafter{\romannumeral #1\relax}}
\usepackage{textcomp,booktabs}

\ifCLASSINFOpdf
\else
\fi
\begin{document}
%
\title{Ontology Based Global and Collective Motion Patterns for Event Classification in Basketball Videos}
%
%
%

\author{Lifang Wu, Zhou Yang, Jiaoyu He, Meng Jian, Yaowen Xu, Dezhong Xu, and Chang Wen Chen

\thanks{This work was supported in part by the Beijing Municipal Education Commission Science and Technology Innovation Project (KZ201610005012), National Natural Science Foundation of China (61702022), China Postdoctoral Science Foundation Funded Project (2017M610026, 2017M610027), Beijing Excellent Young Talent Cultivation Project (2017000020124G075), and Beijing University of Technology "Ri xin" Cultivation Project.

}

\thanks{(Corresponding author: Meng Jian.)}

\thanks{Lifang Wu, Zhou Yang, Jiaoyu He, Meng Jian, Yaowen Xu and Dezhong Xu are with the College of Information and Communication Engineering, Faculty of Information Technology, Beijing University of Technology, Beijing 100124, China, (e-mail: jianmeng648@163.com)}

\thanks{Lifang Wu and Meng Jian are with the Beijing Municipal Key Lab of Computation Intelligence and Intelligent Systems, Beijing University of Technology, Beijing 100124, China}

\thanks{Chang Wen Chen is with School of Science and Engineering, Chinese University of Hong Kong, Shenzhen 518172, China and with Department of Computer Science and Engineering, SUNY Buffalo, NY, USA}}
\maketitle

\begin{abstract}
In multi-person videos, especially team sport videos, a semantic event is usually represented as a confrontation between two teams of players, which can be represented as collective motion. In broadcast basketball videos, specific camera motions are used to present specific events. Therefore, a semantic event in broadcast basketball videos is closely related to both the global motion (camera motion) and the collective motion. A semantic event in basketball videos can be generally divided into three stages: pre-event, event occurrence (event-occ), and post-event. By analyzing the influence of different stages of video segments to semantic events discrimination, it is observed that pre-event and event-occ segments are effective for classification, while post-events are effective for event success/failure classification. In this paper, we propose an ontology-based global and collective motion pattern (On\_GCMP) algorithm for basketball event classification. First, a two-stage GCMP based event classification scheme is proposed. The GCMP is extracted using optical flow. The two-stage scheme progressively combines a five-class event classification algorithm on event-occs and a two-class event classification algorithm on pre-events. Both algorithms utilize sequential convolutional neural networks (CNNs) and long short-term memory (LSTM) networks to extract the spatial and temporal features of GCMP for event classification. Second, we utilize post-event segments to predict success/failure using deep features of images in the video frames (RGB\_DF\_VF) based algorithms. Finally the event classification results and success/failure classification results are integrated to obtain the final results. To evaluate the proposed scheme, we collected a new dataset called NCAA+, which is automatically obtained from the NCAA dataset by extending the fixed length of video clips forward and backward of the corresponding semantic events. The experimental results demonstrate that the proposed scheme achieves the mean average precision of 58.10\% on NCAA+. It is higher by 6.50\% than state-of-the-art on NCAA.

\end{abstract}


%
\IEEEpeerreviewmaketitle

\section{Introduction}
%
%
%
%
\IEEEPARstart{I}{n} recent years, video content analysis has developed rapidly because of the application of deep neural networks [1--8] and large datasets [9--13]. In most surveillance videos [11--15], the main idea of video analysis is to recognize a semantic event based on the action of an object or its interactions. Generally, these events can be presented as one or two persons' actions or crowd activities. Many effective algorithms \cite{6}, [17--19] have been proposed for such video action recognition. In single- or two-person videos, the actions are usually analyzed through key object tracking. In crowd-based videos [20--23], universal properties were extracted for crowd event detection. However, in some application scenarios such as team sports, the number of people is much higher than one or two, but it is also much smaller than that of a crowd. In such videos, object tracking is usually difficult due to the occlusion of objects. Some researchers \cite{6,7} have focused their attention on such problems. In \cite{6,7}, object tracking methods were embedded in the algorithm to mark off the individuals. Ref. \cite{7} further implemented individual action classification by relying on person tractlets. However, in numerous frames, occlusion between players usually brings about tracking error. In this circumstance, person-level context information disturbs the model regression and is possible to cause inter-class misclassification.

\begin{figure}
\centering
\epsfig{file=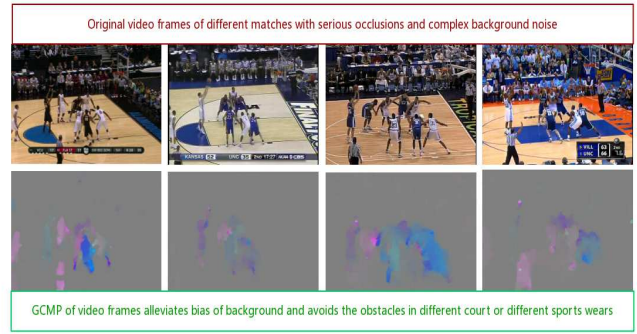,height = 2in,width = 3.5in}
\vskip -0.1in
\caption{Original frames in the first row show the same free throw events in different games with different playing fields, audiences, and shots. The images in the second row are the corresponding GCMP results, which are represented as optical flow. Compared to the original video frames, most of the background noise has been eliminated and the players profiles are already very apparent. Optical flow preserves the location of the players completely and maintains a similar GCMP, thus avoiding the influence of different scenarios.}
\end{figure}

In team sport videos such as basketball \cite{6,38,39,43}, volleyball \cite{7}, and football \cite{37}. Both person level information and group level contexts are crucial cues for event recognition. Some basketball events can be represented by the key individual's motion. Based on this point, Ramanathan et al. \cite{6} detected the key player using an attention model and classifies the event by tracking the key player. They then trained a recurrent neural network model with the local features of each player. However, in some events such as slam duck or steal, the interaction is drastic and some players are often occluded by others. Thus, the key player's motion is not discriminative enough to effectively reflect the features of different events.

From another perspective, a semantic event can be represented by the interaction motions of two groups of people. It can be treated as a typical conflict pattern of two groups of players. To some extent, the conflict pattern can be represented by the motion pattern. Furthermore, in broadcast basketball videos, similar camera motions are utilized to capture videos for the same events. In Fig. 1, a free-throw event from different basketball matches is shown. Obviously, the court color and the team uniforms vary from a match to another. It seems to be a great challenge to make a classifier learn motion patterns from the original spatial domain which contains various changeable factors. Nevertheless, these visual disturbances can be eliminated by translating RGB frames into optical-flows because optical flow focuses on motion instead of color.
Hence, we extract the global and collective motion patterns (GCMPs) utilizing optical flow because the GCMPs have a high similarity for a specific basketball event. Motivated by the above, we propose a GCMP-based event classification algorithm.

In basketball videos, a complete semantic event usually includes three stages, the event preparation (pre-event) stage, event occurrence (event-occ) stage and subsequent actions after the event (post-event) stage. In the pre-event stage, players prepare for the shot. Taking layup for example, the player will hold the ball and move toward the basket in the pre-event. In the event-occ stage, the player will jump up and try to put the ball into the basket. In the post-event stage, the action is finished and the players¡¯ reactions will vary as the shot results. If the shot is failed, the players will struggle for the rebound; otherwise, the players will serve the ball. However, in the NCAA dataset \cite{6}, the video for an event is clipped from the point that the ball is about to leave the shooter's hand to the point at which the ball just comes into contact with the basket. In other words, the videos in the NCAA dataset mostly belong to the event-occ stage of a complete event. For some events, such as free throw, the event-occ stages are discriminative enough for recognition. However, some events (e.g., layup and other two-point) share similar motion patterns in the event-occ stage, but they have distinct motions during the pre-event stage. Furthermore, the features from post-event video segments carry rich information that is closely related to the success (or failure) of the events. Hence, the video clips in the NCAA dataset, which generally include the event-occ video segments only, thus cannot represent the events completely. To cater to the need for a more complete basketball dataset, we collect a new dataset called NCAA+ based on the existing dataset NCAA by extending the fixed length of the video clips forward and backward. Based on NCAA+ dataset and the observations on it, we incorporate ontology into our GCMP scheme for basketball video event classification.

The proposed scheme includes three steps, as shown in Fig. 2. GCMP\_DF\_SVF represents the deep features of GCMP in the sequential video frames while RGB\_DF\_VF means the deep features of images in the video frames. First, we present a two-stage event classification scheme including a GCMP\_DF\_SVF based five-class (three-point, free-throw, layup + other two-point, slam dunk, and steal) event classification algorithm on event-occs and a GCMP\_DF\_SVF based two-class (layup and other two-point) event classification algorithm on pre-events. Both stages utilize the CNN + LSTM structure to extract the spatial and temporal features of GCMP, which is represented as optical flow. Second, we further use post-event segments to predict the success/failure of the event by RGB\_DF\_VF based on the CNN pipeline. Third, the six-class event classification results and the success/failure prediction results are integrated to obtain the final representation of the collective activity. The experimental results demonstrate that the proposed scheme obtains the mean average precision of 58.10\% on NCAA+. It is higher by 6.50\% than Ramanathan's scheme on NCAA \cite{6}. \vspace{3mm}

\begin{figure}
\centering
\epsfig{file=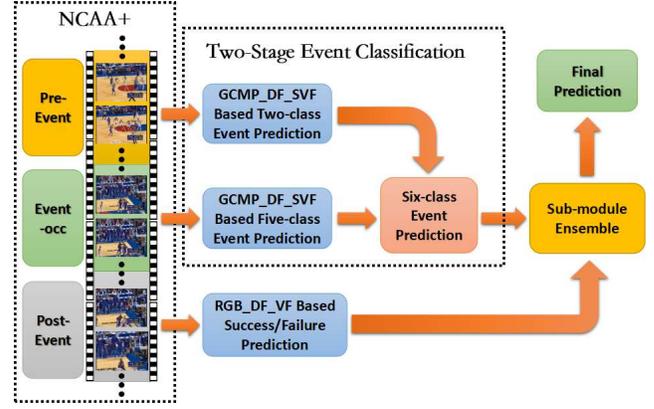,height = 2.1in,width = 3.3in}
\caption{Overview of the proposed On\_GCMP framework for event prediction in basketball videos. Following the ontology manner in basketball games, different event segments is utilized for different objectives. Specifically, basic types of events are firstly predicted by event-occ and pre-event segments and then their success or failure attribute is determined using a parallel network by post-event. Finally, the results of two streams are combined to obtain the final results.}
\end{figure}

The main contributions of this work are summarized as follows.
\begin{itemize}
\item  We introduce the GCMP method for team sport video analysis. This approach globally extracts the motion pattern (and the conflict patterns) of two groups of players as well as the camera motions. \vspace{3mm}
\item We split a semantic event into three stages: pre-event, event-occ, and post-event. We further analyze the correlations between the video segments of different stages and the semantic events. We collect a new dataset called NCAA+, obtained by automatically extending the fixed length of the video segments forward and backward for events in the NCAA dataset \cite{6}. \vspace{3mm}
\item We propose the ontology-based GCMP (On-GCMP) for basketball event classification.
    Different stages of video segments are utilized for different recognition tasks, as shown in Fig. 2. The compared experimental results demonstrate that the proposed scheme on NCAA+ raises the performance of the results obtained by Ramanathan's method on NCAA\cite{6} by 6.50\% on average.

\end{itemize}

%

\section{Related Work}
Video-based human behavior recognition is a topic of intense interest in computer vision research. Here, we briefly overview the related work on these tasks.\vspace{3mm}

\noindent\textbf{1) Single-person action analysis}

Previous work on single action classification utilized various types of handcrafted features to model video sequences such as histogram of oriented gradients (HOG) \cite{10} and histogram of optical flow (HOF) \cite{11} features. More recently, CNN-based methods have achieved state-of-art performance by integrating features in both the spatial and temporal domains. Specifically, Ji et al. \cite{12} proposed a novel three-dimensional (3D) CNN structure called C3D that fuses the feature map from two domains and utilizes 3D filters to merge cues over time and space simultaneously. Wang et al. \cite{9}, Feichtenhofer et al. \cite{13}, and Simonyan et al. \cite{14} designed two-stream CNN structures to learn temporal (optical flow) and spatial (stacked RGB frame) context features separately and integrated the features in the fusion layers. Wang et al. \cite{15} put forward a pyramid network to learn spatial and temporal features jointly and encoded the consecutive information into a robust representation. \vspace{3mm}

\noindent\textbf{2) Collective activity analysis}

Historically, a number of approaches \cite{16,17,18} have treated people in a scenario as 2D points and modelled the collective interaction by structured shapes. The distribution of the 2D points were further utilized for low-order temporal feature encoding. Khan et al. \cite{19} creatively modelled a formation of people as a 3D polygon that is invariant to camera motion to some extent. However, handcrafted feature-based methods cannot sufficiently utilize the interactions among each individual. Recently, many approaches have exploited deep learning pipelines to learn a robust representation of fused features in the spatial and temporal domains. Ramanathan et al. \cite{6} proposed an attention-based model for basketball action recognition: they exploited a recurrent neural network to learn event features and combined attention weights for further improvement. Ibrahim et al. \cite{7} proposed a two-stage hierarchical deep model to integrate person-level features and group level dynamics. Girdhar et al. \cite{42} developed a video-level representation that aggregates convolutional descriptors across different portions of the imaged scene and across the entire temporal span of the video.\vspace{3mm}

\noindent\textbf{3) Crowd activity analysis}

Because occlusions and pose variations are randomly distributed in the crowd scenes, crowd understanding is a challenging topic. Numerous studies on various types of tasks have been conducted in this field such as crowd counting, crowd behavior analysis, and crowd tracking. Kratzand et al. \cite{20} and Mahadevan et al. \cite{21} dealt with anomaly detection tasks using localized motion patterns and statistic distributions to evaluate the global characteristics of a scene. Loy et al. \cite{22} utilized a similar idea, but took advantage of the correlation of different camera views as the local regions. For the crowd counting task, deep learning methods have been widely used to fuse the spatial and temporal semantics. Xiong et al. \cite{23} proposed the convolutional LSTM (ConvLSTM) solution, which captures dependencies over time and space jointly. Sam et al. \cite{24} presented a Switch-CNN structure that merges the crowd density as the latent factor to do the prediction of crowd counting.

\section{NCAA+ dataset}

The NCAA basketball dataset \cite{6} was collected from YouTube videos of different venues taken at different times. The videos involve multi-person games. The dataset includes 11 events: three-point success, three-point failure, two-point success, two-point failure, free-throw success, free-throw failure, layup success, layup failure, slam dunk success, slam dunk failure, and steals. For each event, the start and end points were labeled manually using a crowdsource platform, and the length of an event is about 32 frames on average. In the NCAA dataset, there are a total 250 games about 1.5 hours on average. The total video length is about 375 hours.

The NCAA+ dataset is automatically obtained by extending fixed the length of the video clips forward and backward for the events. It also includes the same 11 events as NCAA. In contrast to NCAA, each video segment in NCAA+ contains 60 frames on average, including about 32 frames of the event-occ, which correspond to the video clips in the NCAA dataset, the preceding 18 frames of the pre-event, and the following 10 frames of the post-event. Figure 3 shows the down-sampled video clips for a free-throw event in the NCAA+ dataset. The frames marked in red borders comprise the pre-event segment, the event-occ frames are marked in blue borders, and the post-event is indicated by green borders.

\begin{figure*}
\centering
\includegraphics[height=0.6\linewidth]{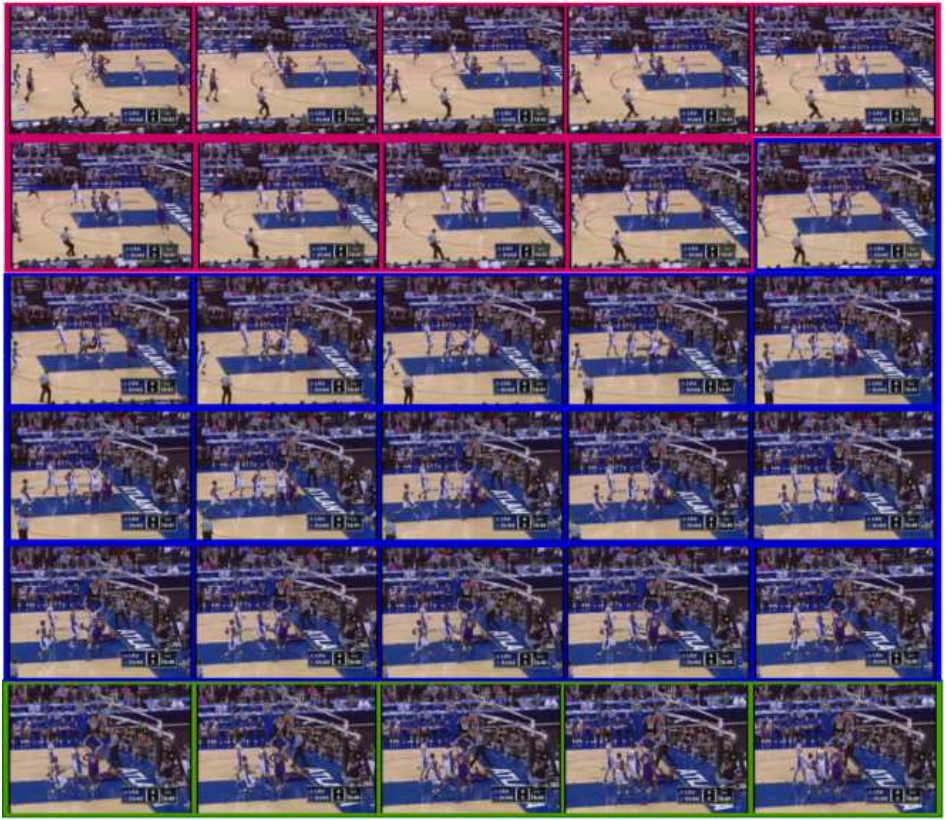}
\caption{Down-sampled example video clip for a free-throw event in the NCAA+ dataset. The pre-event, event-occ, and post-event segments are indicated by red, blue, and green borders, respectively. The event-occ segments correspond to the video clips in the NCAA dataset.}
\end{figure*}

\section{Correlation between video segments of different semantic events}

As described in Section \RNum{3}, a complete semantic event includes three segments: pre-event, event-occ, and post-event segments. First, we analyze the classification ability of different video segments. Then, based on Wu's work \cite{40}, we verify the discrimination of the pre-event and event-occ segments for six events (three-points, free-throws, other two-points, layups, slam dunks, and steals). Finally, we study the effectiveness of the post-event to predict the success/failure of the event. The video segments from a complete match are randomly selected and comprise 13 layups, 8 other two-points, 9 three-points, 8 free-throws, 8 steals, and 3 slam-dunks.

Correlations among different events are computed based on the deep features of images in the sequential video frames (RGB\_DF\_SVF) on event-occ and event-occ + pre-event segments, as shown in Fig. 4. The red bars represent the inter-class correlation while the gray bars represent the intra-class correlation. Fig. 4. shows that the distributions of inter-class and intra-class correlations are highly overlapped. Although the correlations on event-occ + pre-event are slightly more discriminative than those on event-occ, both are indistinguishable for these events. Therefore, the features of RGB\_DF\_SVF are not sufficient for semantic event classification.

\begin{figure}
\centering
\subfigure[]{
\label{Fig.sub.1}
\includegraphics[height = 1.4in,width = 1.6in]{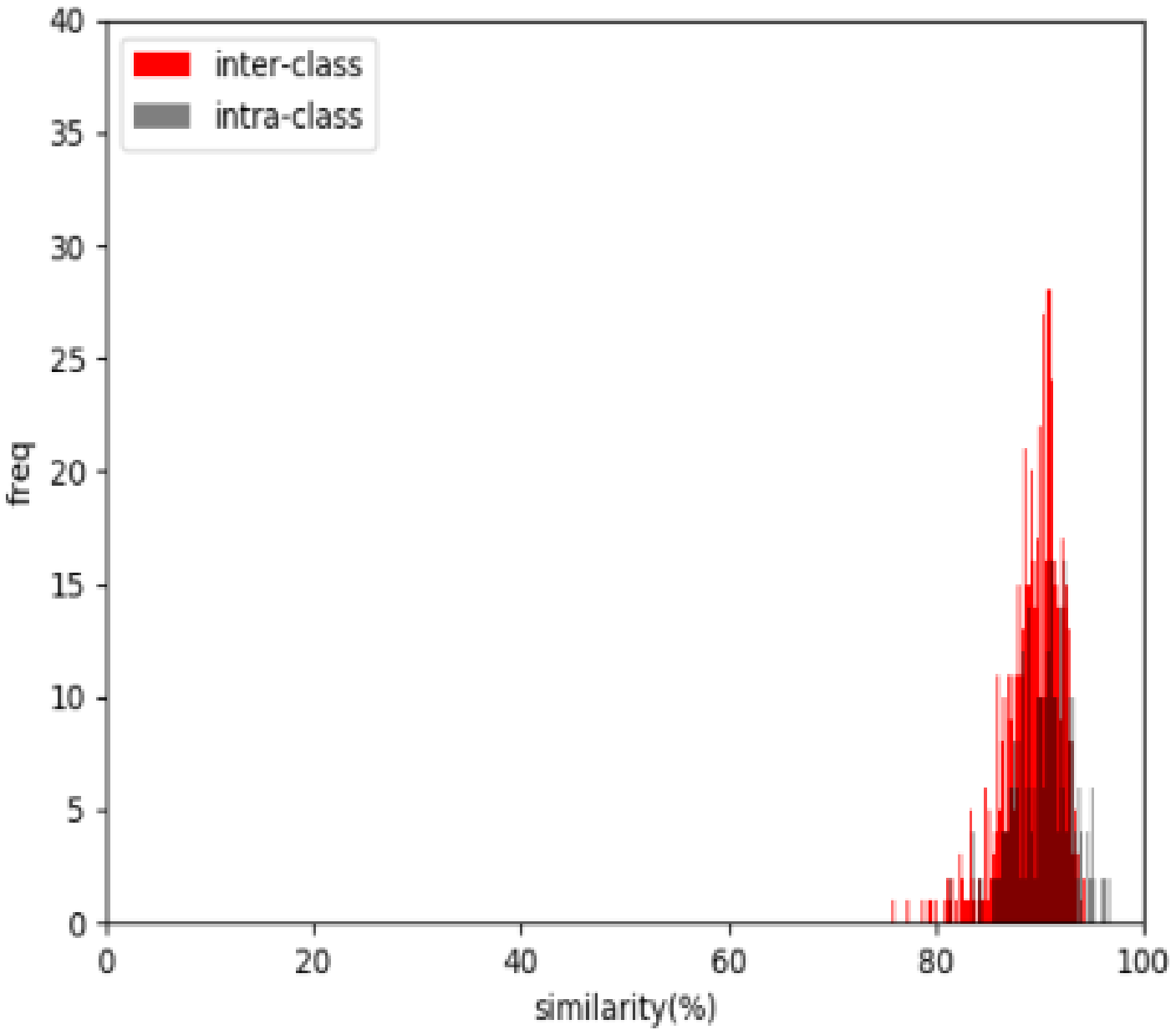}}
\subfigure[]{
\label{Fig.sub.2}
\includegraphics[height = 1.4in,width = 1.6in]{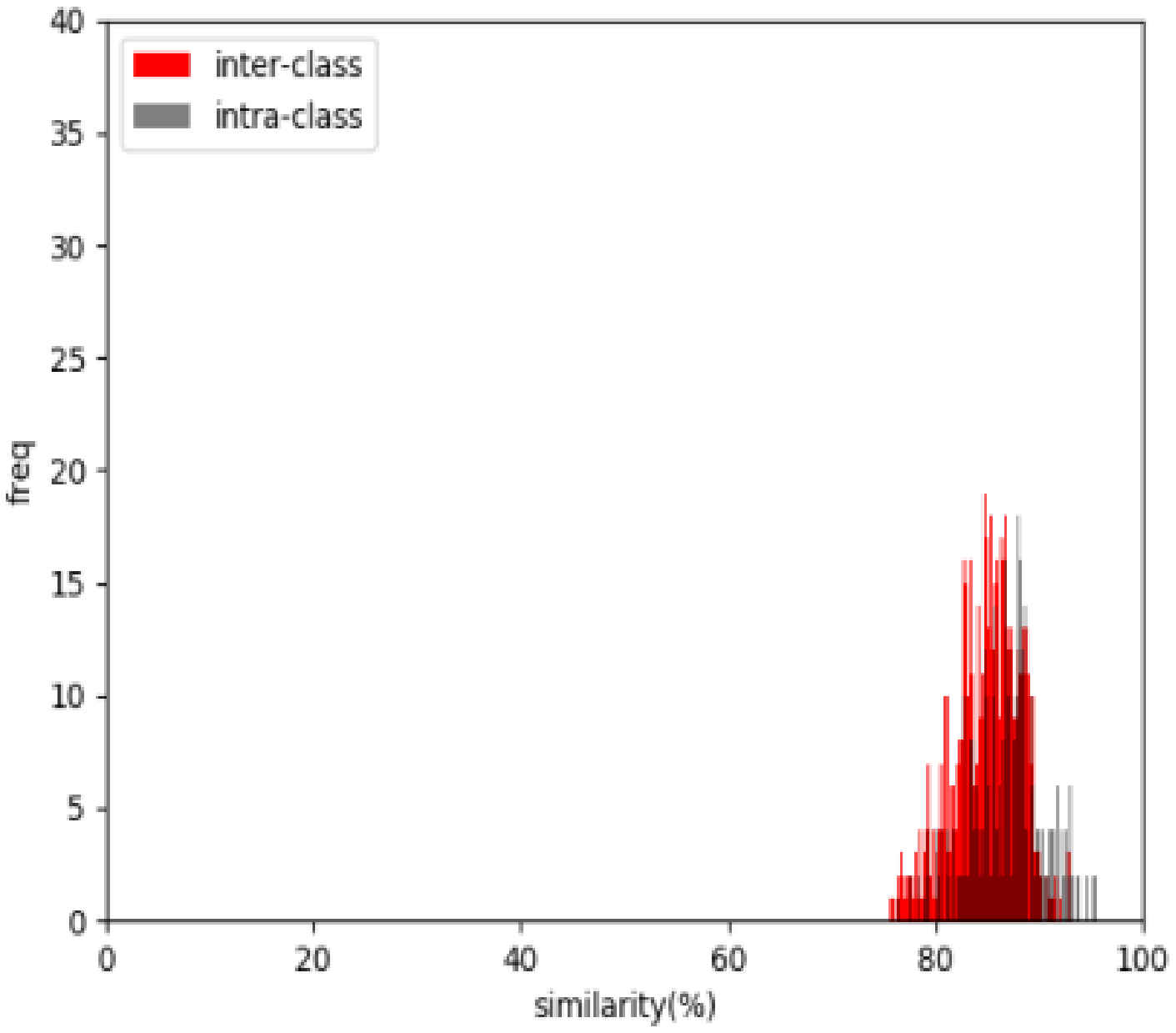}}
\caption{Inter-class and intra-class correlations using RGB\_DF\_SVF features on different video segments. (a) Correlation among event-occ segments. (b) Correlation among event-occ + pre-event.}
\label{Fig.lable}
\end{figure}

We further compute the correlation based on the deep features of GCMP in the sequential video frames (GCMP\_DF\_SVF) on event-occ and event-occ + pre-event segments, as shown in Fig. 5. The inter-class and intra-class correlations are almost distinguishable using the features of GCMP\_DF\_SVF. However, there are still overlaps between the inter-class and intra-class correlations. We further compute the correlations between different class of events for event-occ and event-occ + pre-event segments, as shown in Tables \RNum{1}.

\begin{figure}
\centering
\subfigure[]{
\label{Fig.sub.1}
\includegraphics[height = 1.4in,width = 1.6in]{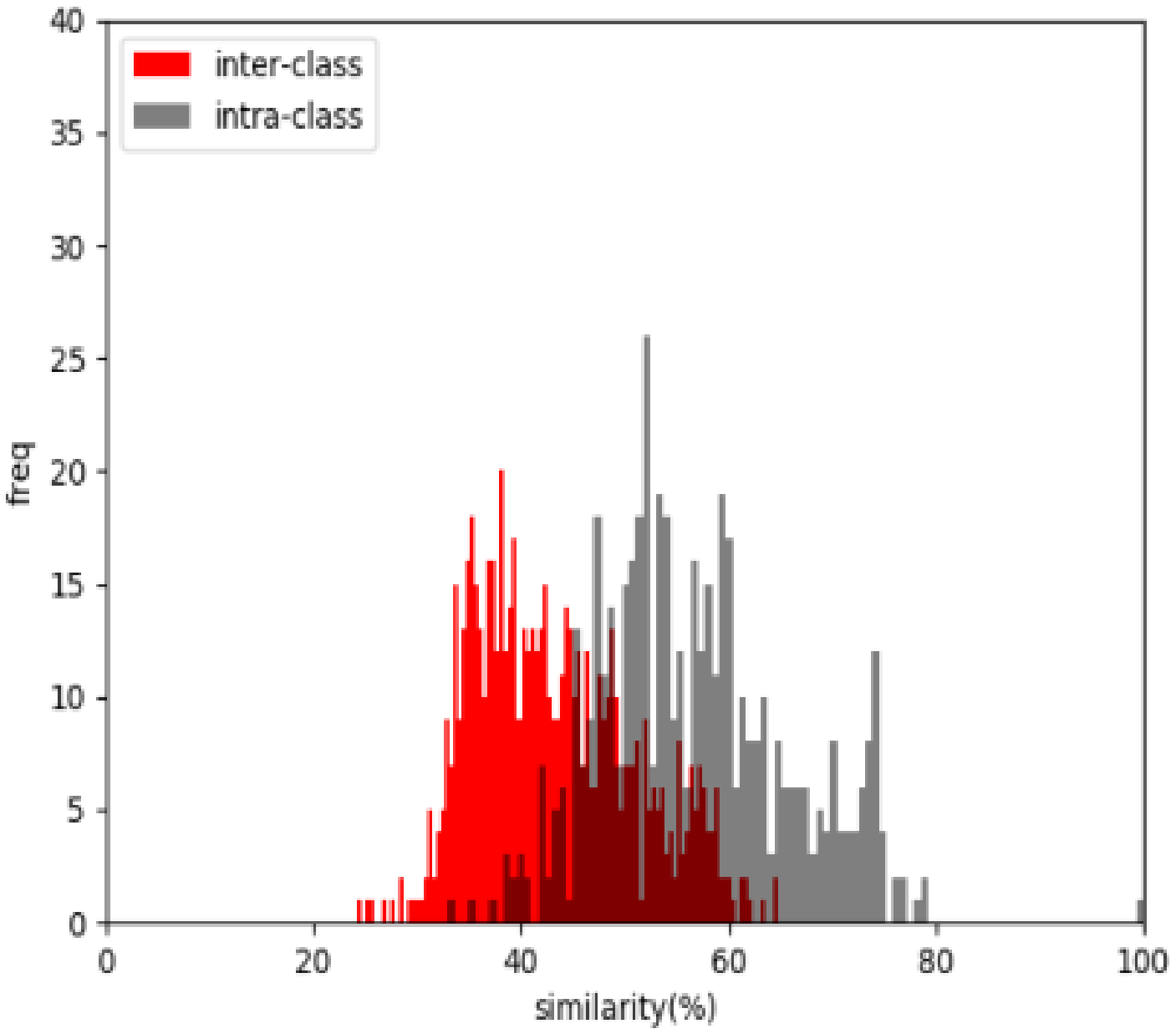}}
\subfigure[]{
\label{Fig.sub.2}
\includegraphics[height = 1.4in,width = 1.6in]{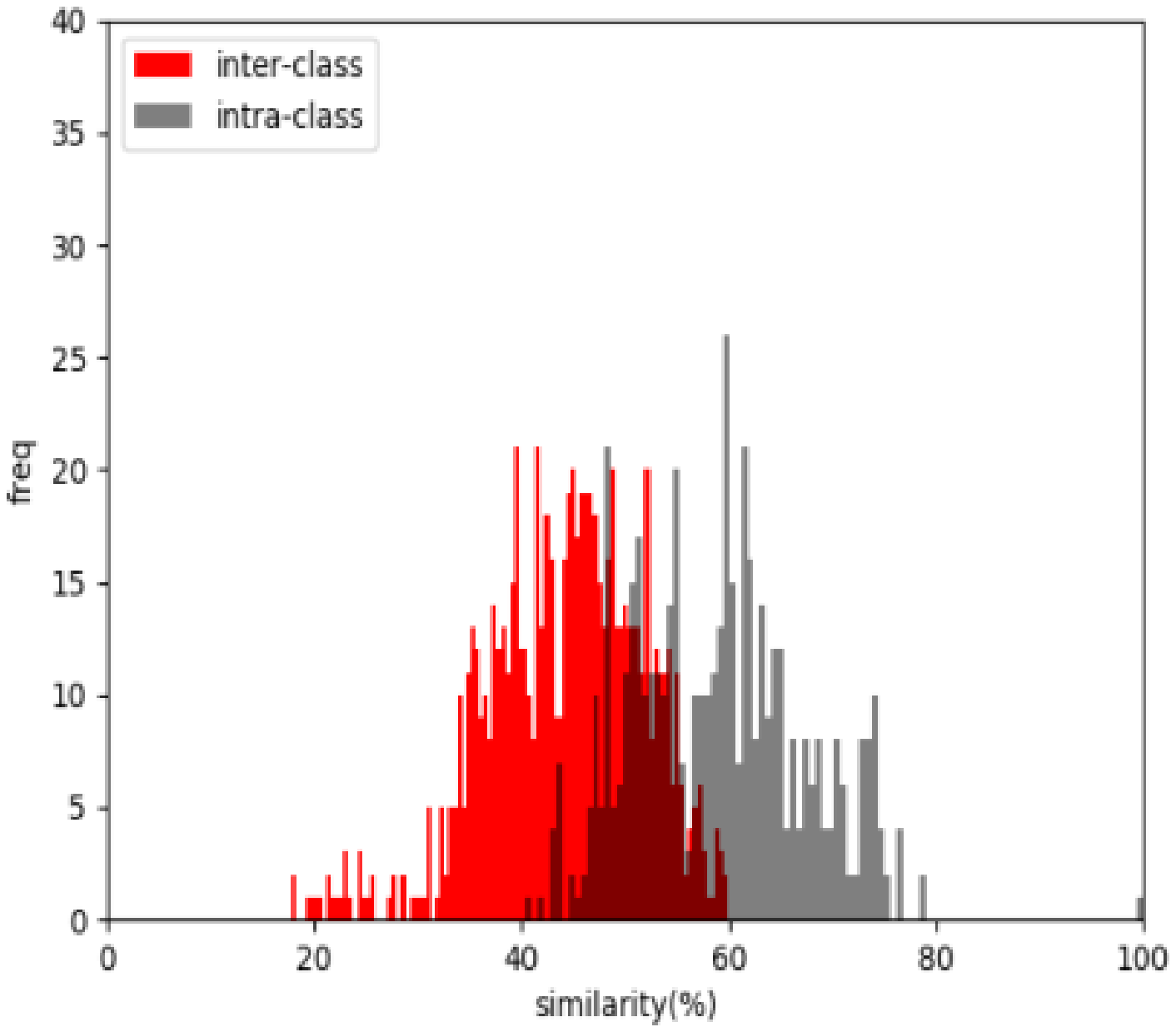}}
\caption{Inter-class and intra-class correlations using the features of GCMP\_DF\_SVF on different video segments. (a) Correlation among event-occ segments. (b) Correlation among event-occ + pre-event segments.}
\label{Fig.lable}
\end{figure}

\makeatletter
\def\hlinew#1{%
  \noalign{\ifnum0=`}\fi\hrule \@height #1 \futurelet
   \reserved@a\@xhline}
\makeatother

\begin{table*}
\caption{Correlation between different events computed by GCMP\_DF\_SVF features on event-occ segments and pre-event + event-occ segments.}
\newcommand{\tabincell}[2]{\begin{tabular}{@{}#1@{}}#2\end{tabular}}
\footnotesize
\centering
\begin{tabular}{|c|c|c|c|c|c|c|c|} 
\hline
\multirow{8}*{\tabincell{c}{Event-occ}}
&Similarity(\%)&Three-point&Free-throw&Layup&Other two-point&Slam dunk&Steal\\  
\cline{2-8}  
&Three-point &60.23&36.61&40.55&43.96&41.22&46.35\\
\cline{2-8}
&Free-throw &36.61&73.53&39.99&42.59&35.47&34.24\\
\cline{2-8}
&Layup &40.55&39.99&52.54&44.91&48.85&45.73 \\
\cline{2-8}
&Other two-point &43.96&42.59&44.91&59.93&44.27&42.89 \\
\cline{2-8}
&Slam dunk &41.22&35.47&48.85&44.27&53.45&49.57 \\
\cline{2-8}
&Steal &46.35&34.24&45.73&42.89&49.57&61.56 \\
\hlinew{2pt}
\multirow{8}*{\tabincell{c}{Pre-event + \\ Event-occ}}
&Similarity(\%)&Three-point&Free-throw&Layup&Other two-point&Slam dunk&Steal\\  
\cline{2-8}  
&Three-point &57.59&22.74&35.49&45.38&41.79&40.19\\
\cline{2-8}
&Free-throw &22.74&75.37&23.24&41.23&20.91&22.47\\
\cline{2-8}
&Layup &35.49&23.24&50.23&42.56&45.79&43.78 \\
\cline{2-8}
&Other two-point &45.38&41.23&42.56&60.21&46.60&36.35 \\
\cline{2-8}
&Slam dunk &41.79&20.91&45.79&46.60&52.14&47.25 \\
\cline{2-8}
&Steal &40.19&22.47&43.78&36.35&47.25&61.24 \\
\hline
\end{tabular}
\end{table*}

From Table \RNum{1}, we can see that the correlations between the similar and different events on event-occ segments are distinguishable for most of events except layup and other two-point events, which are highly correlated with each other. In particular, because of the small amount of slam dunk data, its inter- and intra-class correlations are not as discriminative as other events. At the same time, the correlations of the extended video segments (pre-event + event-occ) decrease a little for both the same class events and different class events. They are discriminative for most events, especially for layup and other two-point events. However, the correlations between two pairs of events (three-point and other two-point, slam dunk and other two-point events) are very high. Therefore, we conclude that the pre-event video segments are effective for discriminating layup and other two-point events but are not effective for three-point/other two-point and slam dunk/other two-point events discriminations.

Hence, we further compute the correlations using the GCMP\_DF\_SVF features on pre-event segments for layup and other two-point events, as shown in Table \RNum{2}. The results show that the layup and other two-point events are distinguishable when pre-event segments are used.

\begin{table}
\footnotesize
\centering
\caption{Correlation between layup and other two-point events using GCMP\_DF\_SVF features on pre-event segments.}
\begin{tabular}{|c|c|c|} \hline
Similarity(\%)&Layup&Other two-point\\ \hline
Layup &49.31&37.17 \\ \hline
Other two-point &37.17&51.41 \\ \hline
\end{tabular}
\end{table}


Considering that success and failure are static states, not dynamic events, we compute the correlation between the success/failure events using the deep features of images in the video frames (RGB\_DF\_VF), the deep features of images in the sequential video frames (RGB\_DF\_SVF), the deep features of images in the video frames (GCMP\_DF\_VF), and the deep features of images in the sequential video frames (GCMP\_DF\_SVF) respectively. The results are shown in Fig. 6. We can see that the success/failure for different events is indistinguishable using the RGB\_DF\_SVF, GCMP\_DF\_VF, and GCMP\_DF\_SVF. However, the RGB\_DF\_VF are effective for success/failure classification.

\begin{figure}
\centering
\subfigure[]{
\label{Fig.sub.1}
\includegraphics[height = 1.4in,width = 1.6in]{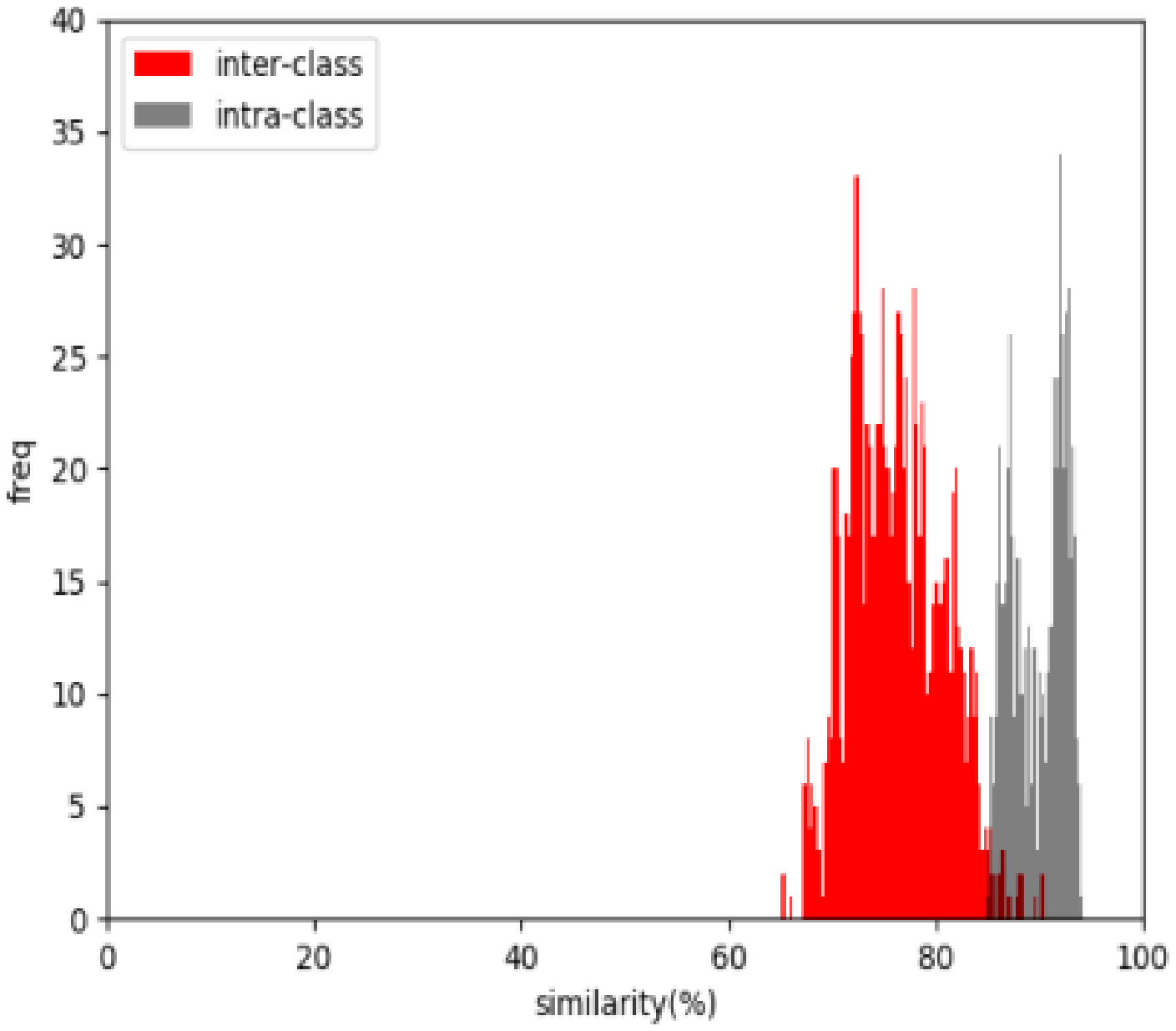}}
\subfigure[]{
\label{Fig.sub.2}
\includegraphics[height = 1.4in,width = 1.6in]{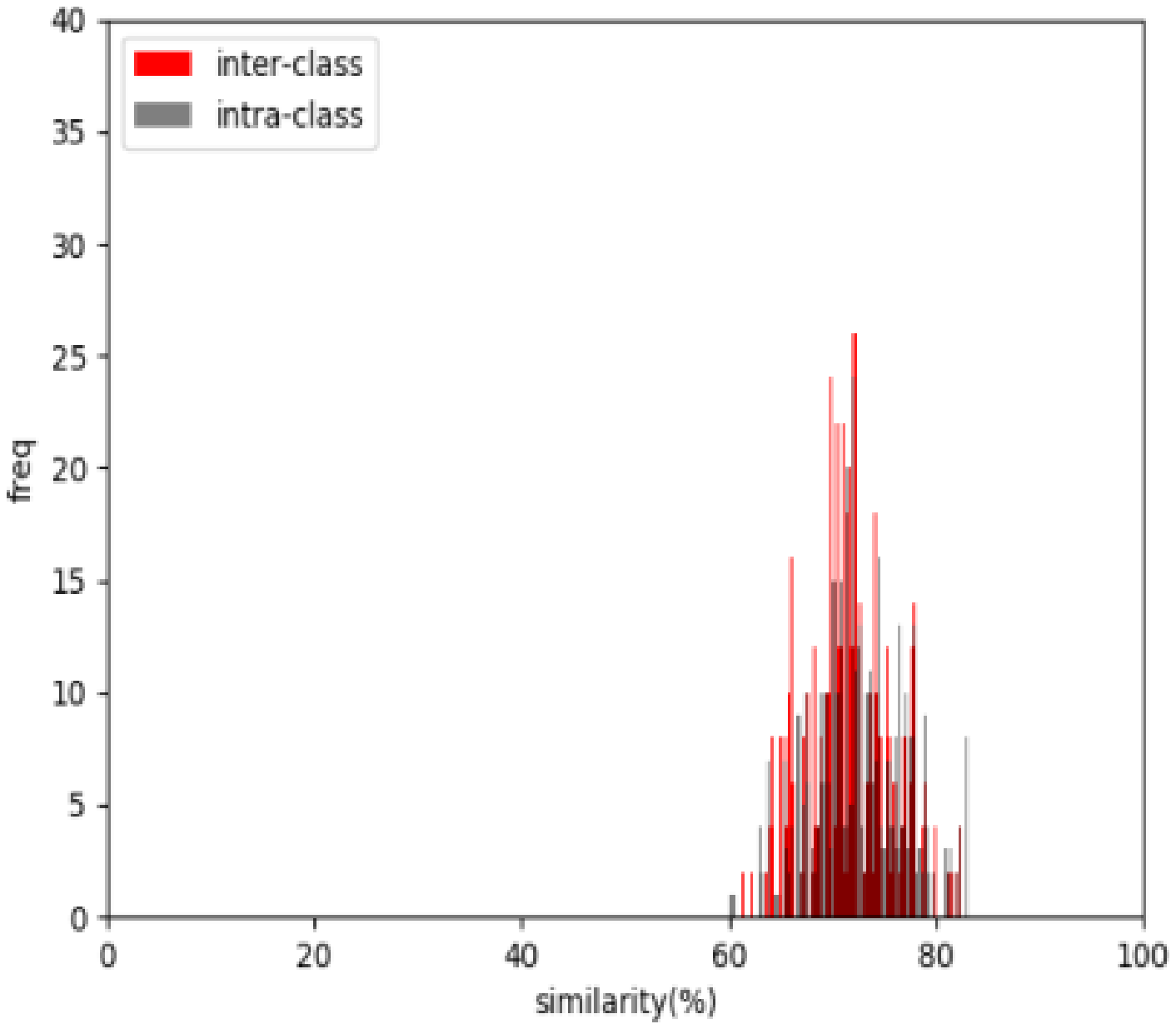}}
\subfigure[]{
\label{Fig.sub.3}
\includegraphics[height = 1.4in,width = 1.6in]{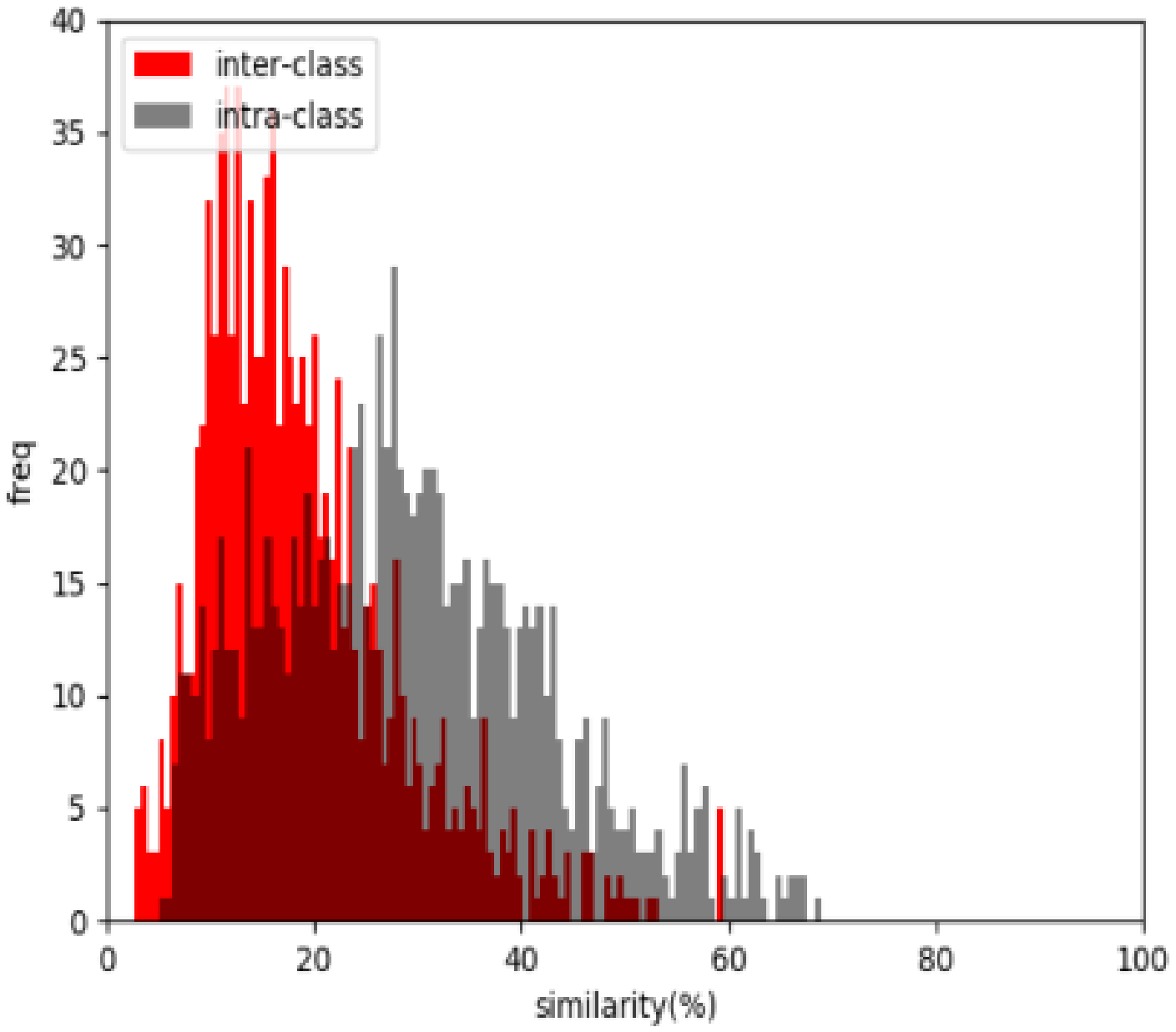}}
\subfigure[]{
\label{Fig.sub.4}
\includegraphics[height = 1.4in,width = 1.6in]{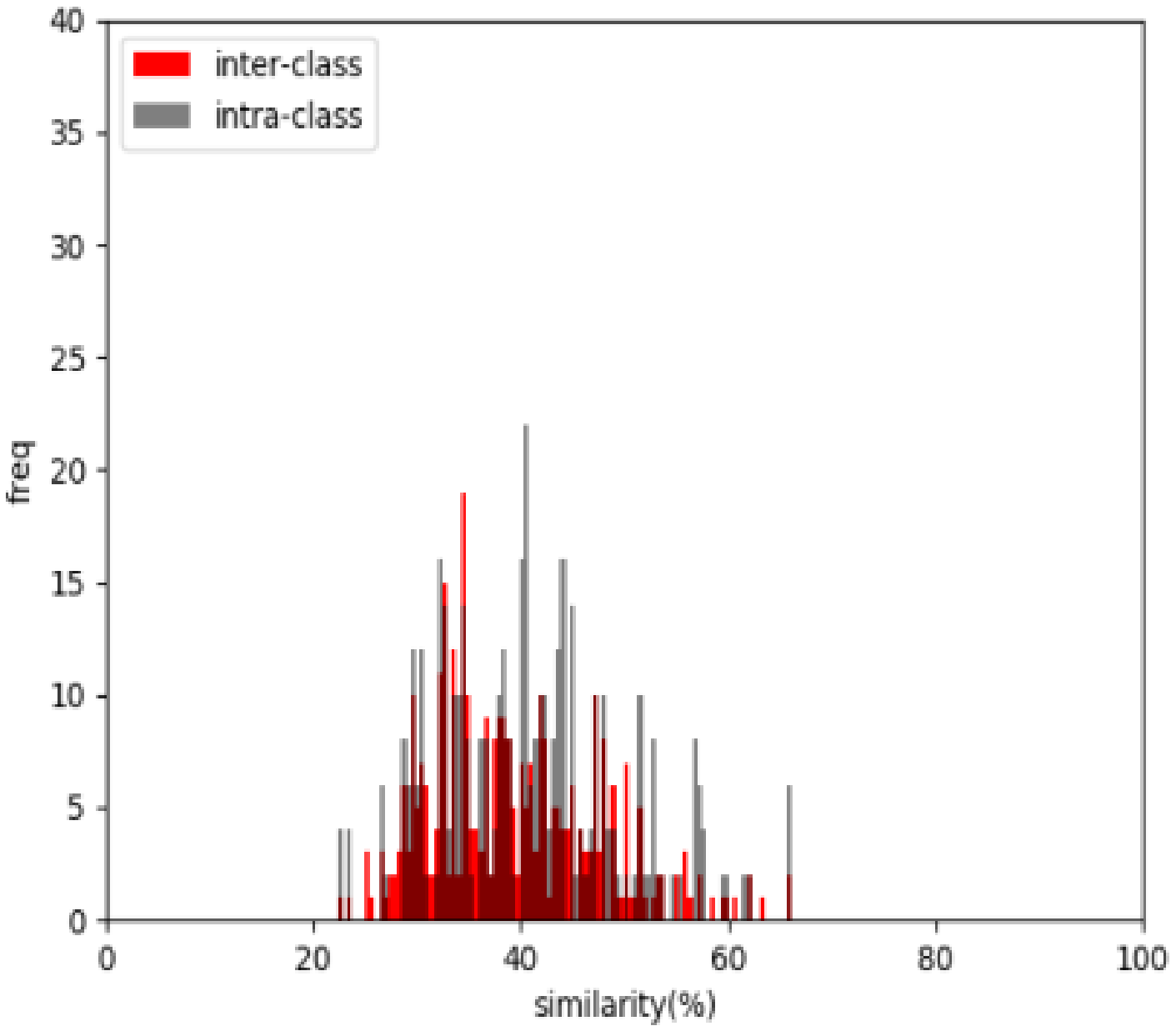}}
\caption{Correlations between post-event segments for success/failure using different deep features: (a) RGB\_DF\_VF, (b) RGB\_DF\_SVF, (c) GCMP\_DF\_VF, and (d) GCMP\_DF\_SVF.}
\label{Fig.lable}
\end{figure}

From the analysis above , we could make the following observations:
\begin{enumerate}
  \item Deep features of GCMP in the sequential images (GCMP\_DF\_SVF) are effective for event classification.
  \item Deep features of the images (RGB\_DF\_VF) are effective for success/failure classification.
  \item Event-occ segments are effective for most events except layup and other two-point events.
  \item Pre-event segments are effective for layup and other two-point classification.
  \item Post-event segments are effective for success/failure classification.

\end{enumerate}

Based on these observations, we propose the ontology based GCMP (On\_GCMP) scheme.

\section{The On\_GCMP based scheme}

\subsection{ Framework of the Proposed On-GCMP Scheme}

According to the observation in Section \RNum{4}, we design an ontology based deep network for basketball event classification. The framework of the proposed scheme is presented in Fig. 2. To integrate ontology manner into our framework, we proposed a two-stage event classification scheme that progressively combines a GCMP\_DF\_SVF based five-event prediction algorithm on event-occ segments and a GCMP\_DF\_SVF based two-event classification algorithm on pre-event segments. In the meantime, a RGB\_DF\_VF based success/failure classification algorithm on post-event segments is designed to discriminate success and failure. Finally, the results of the two-stage event classification scheme and the success/failure classification algorithm are integrated for the overall event prediction.\\

The pipelines can be generalized as follows:
\begin{itemize}
\item \textbf{Two-stage event classification scheme}: The objective of this scheme is to predict the events without considering the property of success or failure. It includes a five-class event classifier and a two-class event classifier. These two classifiers are both designed based on the semantic event classification algorithm using GCMP\_DF\_SVF on video segments. The difference is that the former is trained using event-occ segments of five events (three-point, free-throw, layup + other two-point, slam dunk and steal) while the latter is trained using the pre-event segments of two events (layup and other two-point). In the two-stage scheme, the five-event classifier and the two-event classifier are progressively integrated. For an unknown video segment, its event-occ segment is first input into the five-class classifier, if the output of the classifier is three-point, free-throw, slam dunk and steal, the output is assigned as the event label directly. The two-stage scheme is finished. If the output of five-class classifier is layup + other two-point, the pre-event segment of the video is input into the two-class classifier. The output of the two-class classifier is assigned as the event label.
\item \textbf{Success/Failure classification using RGB\_DF\_VF on post-event segments}: The objective of the algorithm is to determine whether the event is successful or not for all events except steals. Post-event frames are utilized. The features of RGB\_DF\_VF are extracted for success/failure classification. The final results are obtained by voting of the results of all frames. \\
\end{itemize}

From the above two pipelines, we could obtain the six-event prediction vector $V_{6-event}$ = \{3-point, free-throw, layup, other-2-point, slam-dunk, steal\} and success/failure prediction vector $V_{SF}$ = \{succ, fail\} the former 5 elements in $V_{6-event}$ form the five-element vector $V_{5-elem}$ = \{3-point, free-throw, layup, other-2-point, slam-dunk\}. These vectors are all binary vectors. The elements of the predicted event is assigned as '1' and other elements are assigned as '0'.

The Kronecker products of five-element vector $V_{5-elem}$ and success/failure vector $V_{SF}$ outputs the ensemble vector:

\begin{equation}
V_{ensem}=V_{5-elem} \otimes V_{SF}
\end{equation}

where $ \otimes $ represents the Kronecker products operation, $V_{5-elem}$ is a vector including 5 elements, $V_{SF}$ is a vector of 2 elements and $V_{ensem}$ is a vector of 10 elements. The final output of 11-element vector $V_{F}$ could be obtained by connecting $V_{ensem}$ and the sixth element of $V_{6-event}$.

\begin{equation}
\begin{split}
V_{F}=\{V_{ensem}, V_{6-event}[6]\} \\
\end{split}
\end{equation}

\subsection{Semantic Event Classification Using GCMP\_DF\_SVF on Video Segments}
Based on observations in Section \RNum{4}, GCMP\_DF\_SVF are effective for event classification. The five-event classifier and the two-event classifier have the identical structure. In this section, we will introduce the five-event classifier. The architecture of the five-class network can be viewed in Fig. 7. We employ the optical flow to represent the GCMP, because the background information such as the spectators makes little contribution to the classification. CNNs over the spatial domain may extract redundant or irrelevant features, reducing the accuracy of the whole model.

\begin{figure*}
\centering
\includegraphics[height=0.3\linewidth]{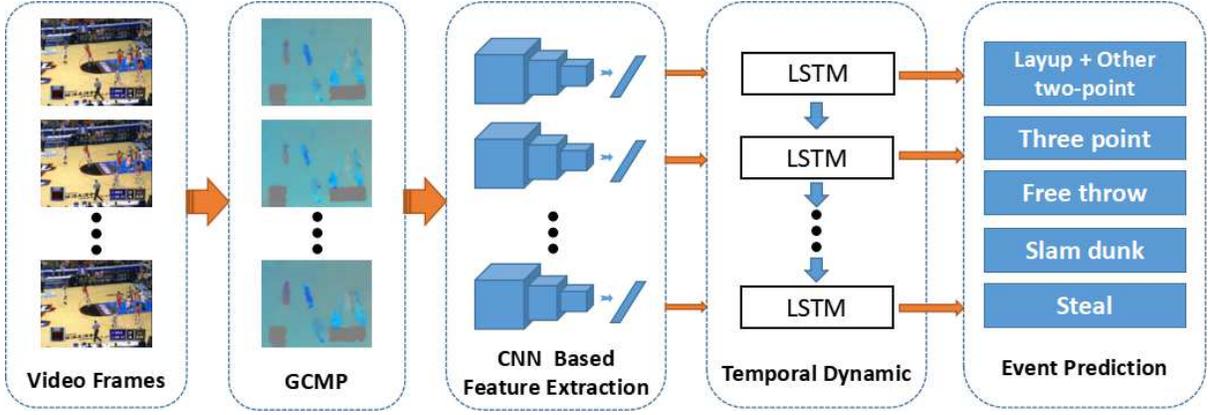}
\caption{Deep network for GCMP\_DF\_SVF based five-class event classification. In this step, layup and other two-point events are integrated as one event and share the same label.}
\end{figure*}

Optical flow is a two-dimensional vector that can express the relative motion between two consecutive frames. In this work, we compute optical flow using the algorithm in  \cite{25} and then convert the results into three-channel images normalized to the range [0,255]. Both motion direction and motion amplitude can be reflected through the color type and shade.

Feature extraction is an essential challenge for many classification task. In \cite{44}, the authors proposed a novel strategy for neutral vector variable decorrelation, which was an important progress in non-Gaussian data feature analysis, such as image and video. In this paper, we utilize another pipeline of feature extraction method and leverage the powerful feature extraction ability of CNN \cite{1} to learn the motion distribution pattern in each optical flow image over the space domain. Then, we adopt an LSTM \cite{27} structure to integrate the spatial features for event prediction.

Recently, LSTM networks \cite{19} have been widely used in numerous video content analysis tasks \cite{28}. LSTM is a special type of recurrent neural network that solves the problem of long-term dependency. The core idea of LSTM is a creative cell unit that is capable of remembering useful data and forgetting redundant data. This function makes LSTM suitable for modeling complex temporal relationships that may span a long range of time. In addition, the spurious gradient updating problem, which often happens in the training state, can be effectively avoided.

Given a video clip with $T+1$ frames ($F_{1}$, $F_{2}$ $\cdots$ $F_{T+1}$), we could obtain optical flow images $o_{1}$, $o_{2}$ $\cdots$ $o_{T}$ from the consecutive frames. Then, we utilize CNN to extract features from each optical flow image. The features are denote using $x_{1}$, $x_{2}$ $\cdots$ $x_{T}$. Then, we input the spatial feature to the LSTM network to extract information over the temporal domain.

In this paper, we employ a single layer LSTM for temporal feature extraction and event prediction. Assuming $x_{t}$ is the feature generated by the CNN, it is input to the LSTM cell at time $t$. Moreover, $h_{t}$ is a 256 dimension vector which denotes the hidden layer output at time $t$.

\begin{equation}
\left\{
             \begin{array}{lr}
             h_{1}=F_{w}(x_{1},h_{0})=F_{w}(x_{1},0) & \\
             h_{2}=F_{w}(x_{2},h_{1}) & \\
             \vdots & \\
             h_{T}=F_{w}(x_{T},h_{T-1}) &
             \end{array}
\right.
\end{equation}

Assume that $N$ is the number of class. For the t$^{th}$ frame, the response value $s_{tn}$ of the corresponding n$^{th}$ neuron $(n \in \{1,2,...N\})$ in the classification layer can be represented as:


\begin{equation}
s_{tn}=\sum_{i=1}^{256} h_{ti}\ast w_{in}+ b_{n}
\end{equation}

For the prediction result, the softmax function is implemented to compute the category with the maximum probability. Assume that $p_{tn}$ is the probability value of the n$^{th}$ class and it can be expressed as follows:

\begin{equation}
p_{tn}=\frac{exp(s_{tn})}{\sum_{n=1}^{N} exp(s_{tn})}
\end{equation}


\begin{equation}
{G}= \frac {\sum_{t=1}^{T} \{ {p}_{t1}, {p}_{t2}...{p}_{tN} \} } {T}
\end{equation}

where $G$ denotes the final prediction vector of the input video clip and $\{{p}_{t1}, {p}_{t2}...{p}_{tn}\}$ is the score vector of the t$^{th}$ frame. From $G$, we can obtain five-event and two-event binary prediction results.

\subsection{Success/Failure Classification Using RGB\_DF\_VF on Post-Event Video Segments}
In this section, post-event video segments are utilized for success/failure prediction. We propose a CNN structure to extract spatial features on the post-event segments for this binary-class task. Although optical flow conveys more motion patterns than spatial images, the spatial distribution of the players between two classes is more distinct in the post-stage, as mentioned in Section \RNum{4}. Moreover, in addition to the player reactions and trajectory of the ball, the basket is also an important reference. Because the basket is stationary, optical flow is incapable of representing the relationship between the basket and ball. Hence, we utilize the features extracted by CNN in the post-event segments. The overview of the structure is shown in Fig. 8. Each CNN pipeline outputs a predicted value, and the final results are obtained by voting of all prediction results.

\begin{figure*}
\centering
\epsfig{file=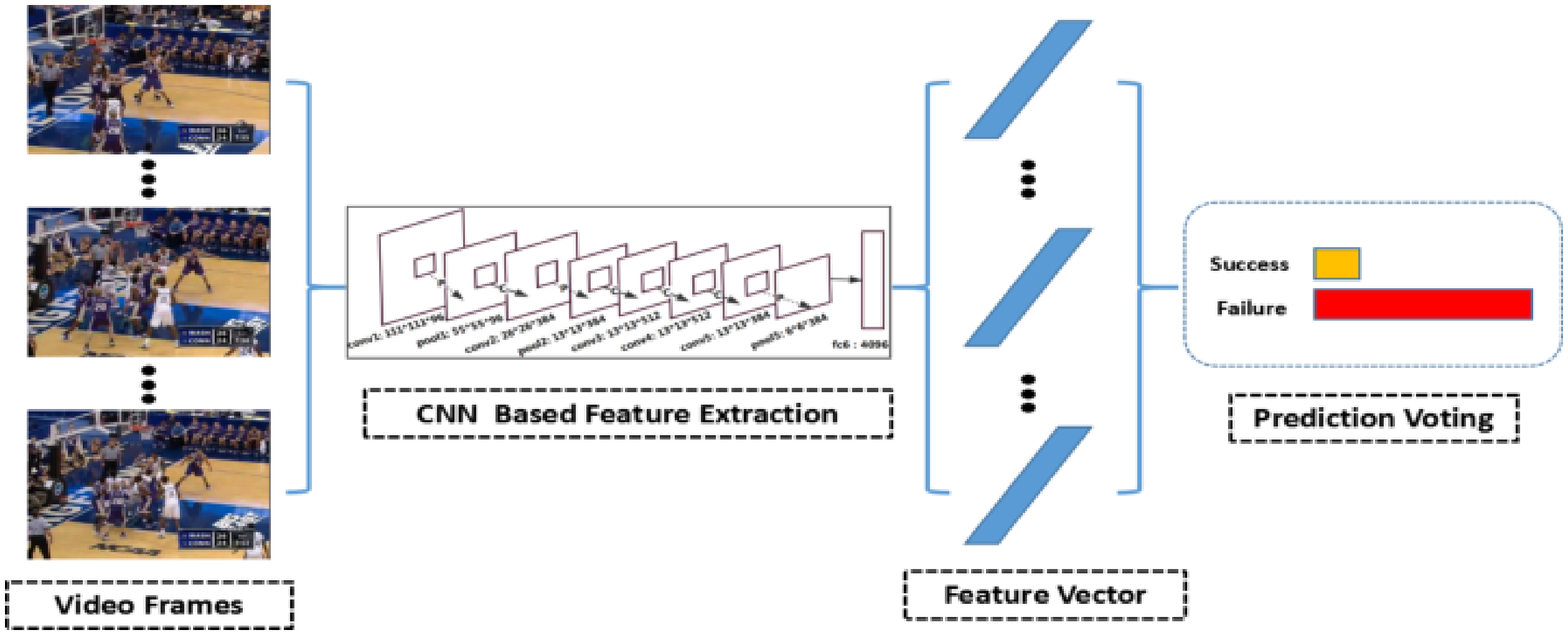,height = 2.0in,width = 5in}
\caption{Architecture of the RGB\_DF\_VF based success/failure prediction network.}
\end{figure*}

\section{Experiments}

In this section, we conduct the experiments on NCAA[11], NCAA+ and a dataset collected in NBA and CBA videos. We evaluate the effectiveness of GCMP and ontology. We also test the performance of the two-stage event classification scheme and success/failure classification algorithm. Furthermore, we compare the proposed On\_GCMP scheme with the state-of-the-art algorithms. Finally, we test the generalization ability of the proposed scheme.

\subsection{Dataset}

We automatically obtain NCAA+ dataset from NCAA by prolonging the fixed length forward and backward for the video clips of the semantic events. Therefore, the semantic events in NCAA+ datasets is identical with those in NCAA dataset. Referring to Ramanathan's experiments[11], the videos of total 250 games are randomly divided into training set (including 200 games) and testing set I (including 50 games). The training set contains video segments of 9,407 semantic events. And the testing set I contains video segments of 2,279 semantic events. The semantic events in both training set and the testing set I includes 11 event categories, which are three-point success and fail, free-throw success and fail, layup success and fail, other two-point success and fail, slam dunk success and fail, and steal. Furthermore, the testing set I is composed of 547 three-points (success and fail), 157 free-throws (success and fail), 428 layups (success and fail), 671 other two-points (success and fail), 31 slam dunks (success and fail) and 445 steals.

\begin{table}
\centering
\footnotesize
\caption{Number of events in Testing set \RNum{2}}
\begin{tabular}{|c|c|c|c|} \hline
Event&From NBA&From CBA&Total\\ \hline
Three-point succ. &19&19&38 \\ \hline
Three-point fail. &27&30&57 \\ \hline
Free-throw succ. &5&3&8 \\ \hline
Free-throw fail. &2&6&8 \\ \hline
Layup succ. &21&24&45 \\ \hline
Layup fail. &4&8&12 \\ \hline
Other two-point succ. &30&22&52 \\ \hline
Other two-point fail. &24&30&54 \\ \hline
Slam dunk succ. &3&2&5 \\ \hline
Slam dunk fail. &2&1&3 \\ \hline
Steal &5&6&11 \\ \hline
Total &142&151&293 \\ \hline
\end{tabular}
\end{table}

Besides the testing set I, we further collect the testing set II from two basketball game videos of National Basketball Association(NBA) and Chinese Basketball Association(CBA) respectively. In the testing set II, We annotate the ¡°start-point¡± and the ¡°end-point¡± of the new test data in the same way as NCAA dataset and extend the video clips forward and backward following the rules of NCAA+ dataset. The number of each event is listed in Table \RNum{3}.

\subsection{Implementation Details}
The implementation of the different methods is described in detail as follows.\\

\begin{itemize}
  \item \textbf{Five-event classification based on event-occ}
\end{itemize}

  1) \textbf{CNN}: In this method, it is necessary to pre-train a CNN model based on video frames. We implement the CNN model using Caffe \cite{41} and the final model is achieved by fine-tuning AlexNet \cite{1}, which extracts high-level semantic features from images. During the CNN training and testing period, we randomly select 15,443 video frames for training and 3,341 video frames for validation. We use a batch size of 128 and a learning rate of 0.001, which is reduced by a factor of 0.95 every 5,000 iterations. The last fully-connected layer has five neurons corresponding to the five categories of three-point, free-throw, layup+other two point, slam dunk, and steal events. To evaluate the ability of the CNN to extract the spatial features for event recognition, we perform event recognition using CNN with frames for 2,279 testing clips.

  2) \textbf{GCMP-based CNN}: The training phase of this method is similar to that of  \textbf{CNN} training, and the training parameters are the same as those of \textbf{CNN}. The difference is that the input frames are optical flow images, which greatly reduces the bias caused by background noise in the video frames and maintains a stable GCMP.

  3) \textbf{CNN+LSTM}: To analyze the effect of LSTM in event recognition, we construct a CNN+LSTM network by fine-tuning the trained CNN, which consists of a CNN connected with a single-layer LSTM. There are 256 hidden nodes in the LSTM layer. Similar to the CNN method for event recognition, a softmax layer is also deployed as a classification layer that corresponds to the five categories of events. Considering the sample imbalance of the five event categories, during the training phase of the LSTM, we randomly sample 4,899 clips for training and 2,279 clips for testing. In this study, we train the model of CNN+LSTM with 16 consecutive frames together for a video. If the last intercept exceeds the end of the video, we take the last 16 frames from the video.

  4) \textbf{GCMP based CNN+LSTM}: The frames are preprocessed by optical flow and fed to \textbf{CNN + LSTM} to extract spatial and temporal features for event recognition. The training parameters are the same as those of \textbf{CNN + LSTM}.\\

\begin{itemize}
  \item \textbf{Two-event classification based on pre-event segments}
\end{itemize}

  The implementation of this algorithm is similar to that of five-event classification algorithm based on event-occ segments. The difference is that the last fully connected layer has 2 neurons corresponding to layup and other two-point events. And the input video clips are pre-event segments. We randomly select 2000 video clips from pre-event for training, including 1000 layups and 1000 other two-point events. Then, the other 1,099 video segments in the testing dataset are utilized for testing. Other parameters are set the same as the five-event classification experiment. \\

\begin{table}
\newcommand{\tabincell}[2]{\begin{tabular}{@{}#1@{}}#2\end{tabular}}
\centering
\footnotesize
\caption{Comparison of event classification performance with/without GCMP on event-occ segments.}
\begin{tabular}{|c|c|c|} \hline
Accuracy(\%)& Without GCMP & With GCMP \\ \hline
Three-point &69.32 &68.56 \\ \hline
Free-throw &20.86  &92.99 \\ \hline
Layup+other two-point &72.22 &74.98 \\ \hline
Slam dunk &5.36 &16.13 \\ \hline
Steal &80.51 &87.87 \\ \hline
Average(\%) &49.65 &68.11  \\ \hline
\end{tabular}
\end{table}

\begin{table*}
\centering
\footnotesize
\caption{Confusion matrix for GCMP\_DF\_SVF based event classification on event-occ segments.}

\begin{tabular}{|c|c|c|c|c|c|c|c|} \hline 
\diagbox{Groundtruth}{Prediction}&Three-point&Free-throw&Layup&Other two-point&Slam dunk&Steal&Accuracy(\%)\\ \hline
Three-point &\textbf{347}&13&8&119&6&54&63.44 \\ \hline
Free-throw &3&\textbf{150}&0&0&0&4&95.54 \\ \hline
Layup &43&13&\textbf{98}&152&26&96&22.90 \\ \hline
Other two-point &126&29&49&\textbf{333}&11&123&49.63 \\ \hline
Slam dunk &19&1&0&1&\textbf{6}&4&19.35 \\ \hline
Steal &1&5&0&1&0&\textbf{438}&98.43 \\ \hline
Average(\%) &\verb|--|&\verb|--|&\verb|--|&\verb|--|&\verb|--|&\verb|--|&58.22 \\ \hline
\end{tabular}
\end{table*}

\begin{table*}
\centering
\footnotesize
\caption{Confusion matrix for the five-class event classification from event-occ segments.}
\begin{tabular}{|c|c|c|c|c|c|c|} \hline
\diagbox{Groundtruth}{Prediction}&Three-point&Free-throw&Layup and Other two-point&Slam dunk&Steal&Accuracy(\%)\\ \hline
Three-point &375&0&99&1&72&68.56 \\ \hline
Free-throw &3&146&4&0&4&92.99 \\ \hline
Layup and other two-point &68&28&824&11&168&74.98 \\ \hline
Slam dunk &19&1&1&5&4&16.13 \\ \hline
Steal &11&5&38&0&391&87.87 \\ \hline
Average(\%) &\verb|--|&\verb|--|&\verb|--|&\verb|--|&\verb|--|&68.11 \\ \hline
\end{tabular}
\end{table*}

\begin{itemize}
  \item \textbf{Success/Failure classification based on post-event segments}
\end{itemize}

  In order to verify the ability of the original frame-based post-event to classify success and failure, we adopt a method similar to the CNN for event classification based on event-occ segments. During the training and testing period, we randomly select 7,383 video frames from post-events for training and 2,279 video frames for testing.

\subsection{Effectiveness of Global and Collective Motion Patterns (GCMP)}

This experiment is designed to evaluate the effectiveness of GCMP. In this experiment, we implement the framework in Fig. 7. for five semantic events (three-point, free-throw, layup + other two-point, slam dunk, and steal) in two ways. One way is the complete framework in Fig. 7. The other way is to skip the GCMP extraction step, and the video frames are input into the CNN-based feature extraction module directly. The compared results are shown in Table \RNum{4}.

From Table \RNum{4} we can see that by introducing GCMP, the accuracy of free three, layup + other two-point, slam dunk and steal  absolutely increases by 72.13\%,12.76\%,10.83\% and 27.36\%, while the accuracy of three-point absolutely decreases a little by 0.76\%. Anyway, the average accuracy could increase by 18.46\%. Therefore, we could say that the GCMP is effective for semantic event classification.


\begin{table}
\centering
\footnotesize
\caption{Prediction results of layup and other two-point events on pre-event segments.}
\begin{tabular}{|c|c|c|c|} \hline
&Layup&Other two-point&Average\\ \hline
Test clips &386&438&\verb|--| \\ \hline
Successfully predict &189&376&\verb|--| \\ \hline
Accuracy(\%) &48.96&85.81&67.39 \\ \hline
\end{tabular}
\end{table}

\begin{table}
\centering
\footnotesize
\caption{Six-event prediction accuracy using the two-stage event classification scheme.}
\begin{tabular}{|c|c|c|c|} \hline
Event&Test set&Successfully predict&Accuracy(\%)\\ \hline
Three-point &547&375&68.56 \\ \hline
Free-throw &157&146&92.99 \\ \hline
Layup &428&189&\textbf{44.16} \\ \hline
Other two-point &671&376&\textbf{56.04} \\ \hline
Slam dunk &31&5&16.13 \\ \hline
Steal &445&391&87.87 \\ \hline
Average &\verb|--|&\verb|--|&60.96 \\ \hline
\end{tabular}
\end{table}

\subsection{Effectiveness of the two-stage event classification scheme}
According to the observations in Section \RNum{4}, there is relatively small correlation among all the events except for layup and other two-point events on Event-occ segments. To verify it, we directly extract GCMP\_DF\_SVF features from event-occ segments for event classification (six-class events). The confusion matrices are shown in Table \RNum{5}.

In Table \RNum{5}, the average prediction accuracy is 58.22\%, three-point, free-throw, and steal events obtain an accuracy of over 60\%, while layup and slam-dunk events have an accuracy of around 20\%. The fourth row in Table \RNum{5} shows that the video clips of 35\% layup are falsely classified as other two-point events.

We then test the performance of the proposed two-stage event classification algorithm. First, we extract GCMP\_DF\_SVF features from event-occ segments for five-class event classification. The confusion matrices are shown in Table \RNum{6}, which shows that the average accuracy is increased by about 10\% with respect to the results in Table \RNum{5}. These results are in accordance with the observations in Section \RNum{4}. We then extract GCMP\_DF\_SVF features from pre-event segments for two-class event classification (layup and other two-point events). The results are shown in Table \RNum{7}. In this step, the layup and other two-point events obtain 48.96\% and 85.81\% accuracy, respectively. By combining the results in two steps, we obtain the final results of the two-stage algorithm, as shown in Table \RNum{8}.

\begin{table}
\newcommand{\tabincell}[2]{\begin{tabular}{@{}#1@{}}#2\end{tabular}}
\centering
\scriptsize
\caption{Performance comparison of feature extraction schemes on the event success/failure prediction task.}
\begin{tabular}{|c|c|c|c|c|} \hline
Model&\tabincell{c}{RGB\_DF\\\_VF}&\tabincell{c}{GCMP\_DF\\\_VF}&\tabincell{c}{RGB\_DF\\\_SVF}&\tabincell{c}{GCMP\_DF\\\_SVF} \\ \hline
\tabincell{c}{Accuracy \\ (\%)} &\textbf{99.78}&71.15&74.03&77.74 \\ \hline
\end{tabular}
\end{table}

\begin{table}
\newcommand{\tabincell}[2]{\begin{tabular}{@{}#1@{}}#2\end{tabular}}
\centering
\scriptsize
\caption{Comparison of performance with/without ontology on NCAA+.}
\begin{tabular}{|c|c|c|c|} \hline
Accuracy(\%)
&\tabincell{c}{GCMP+\\CNN+LSTM}
&\tabincell{c}{GCMP+CNN+\\LSTM and CNN}
&\tabincell{c}{ \ On\_GCMP\ } \\ \hline
Three-point succ. &26.58&51.28&65.91 \\ \hline
Three-point fail. &53.01&52.91&70.34 \\ \hline
Free-throw succ. &86.99&89.55&92.54 \\ \hline
Free-throw fail. &21.33&83.33&93.33 \\ \hline
Layup succ. &61.67&18.99&44.69 \\ \hline
Layup fail. &1.92&20.88&50.60 \\ \hline
Other two-point succ. &20.10&46.79&58.51 \\ \hline
Other two-point fail. &73.08&40.41&54.48 \\ \hline
Slam dunk succ. &0&9.09&18.18 \\ \hline
Slam dunk fail. &0&25&15 \\ \hline
Steal &86.12&98.43&87.87 \\ \hline
Average(\%) &39.16&48.48&59.22 \\ \hline
\end{tabular}
\end{table}

Table \RNum{8} shows that the accuracy of two-stage event classification scheme has increased by 21.26\% and 6.41\% for layup and other two-point events respectively, compared with that in Table \RNum{5}. In the meantime, the accuracy of other events has changed by 5.12\%, -2.55\%, -3.22\%, and -10.56\% respectively. Moreover, the average accuracy has been increased by 2.74\%. These results show that the two-stage classification scheme is effective.

\begin{table}
\newcommand{\tabincell}[2]{\begin{tabular}{@{}#1@{}}#2\end{tabular}}
\centering
\footnotesize
\caption{Performance of different algorithms.}
\begin{tabular}{|c|c|c|} \hline
MAP(\%)
&\tabincell{c}{Ramanathan \cite{6}\\(NCAA)}
&\tabincell{c}{ \ On\_GCMP\  \\ \ (NCAA+)} \\ \hline
Three-point succ. &60&\textbf{73.68} \\ \hline
Three-point fail. &73.80&\textbf{75.28} \\ \hline
Free-throw succ. &\textbf{88.20}&65.59 \\ \hline
Free-throw fail. &51.60&\textbf{77.23} \\ \hline
Layup succ. &\textbf{50}&49.13 \\ \hline
Layup fail. &44.50&\textbf{52.83} \\ \hline
Other two-point succ. &34.10&\textbf{62.93} \\ \hline
Other two-point fail. &47.10&\textbf{65.47} \\ \hline
Slam dunk succ. &29.10&\textbf{30.77} \\ \hline
Slam dunk fail. &0.4&\textbf{25} \\ \hline
Steal &\textbf{89.30}&61.19 \\ \hline
Average(\%) &51.60&\textbf{58.10} \\ \hline
\end{tabular}
\end{table}

\begin{table*}
\newcommand{\tabincell}[2]{\begin{tabular}{@{}#1@{}}#2\end{tabular}}
\centering
\footnotesize
\caption{Performance of two-stage event classification scheme on testing set I and II.}
\begin{tabular}{|c|c|c|c|c|c|c|c|} \hline
--&\--&\
Three-point&\
Free-throw&\
Other two-point&\
Layup&\
Slam dunk&\
Steal\\ \hline
\multirow{2}*{\tabincell{c}{Testing set II}}
&Number of event&95&16&106&57&8&11 \\
\cline{2-8}
&Accuracy(\%)&56.84&0&52.83&49.12&0&90.91 \\ \hline
\multirow{2}*{\tabincell{c}{Testing set I}}
&Number of event&547&157&671&428&31&445 \\
\cline{2-8}
&Accuracy(\%)&68.56&92.99&56.04&44.16&16.13&87.87 \\ \hline
\end{tabular}
\end{table*}

\subsection{Effectiveness of the success/failure classification algorithm}
We compare the performance using different models on post-event segments in Table \RNum{9}, inluding RGB\_DF\_VF, GCMP\_DF\_VF, RGB\_DF\_SVF and GCMP\_DF\_SVF. Table \RNum{9} shows that GCMP does not improve the prediction performance because the motion pattern is ambiguous in this stage. In addition, the LSTM structure substantially decreases the accuracy from 99.87\% to 74.03\%. The basket in a post-event segment is a crucial reference; thus, static spatial features RGB\_DF\_VF gets the best performance. These results are in accordance with the observations in Section \RNum{4}.

\subsection{Effectiveness of the ontology}
In order to evaluate the performance of the proposed ontology, we conduct the compared experiments on the same dataset NCAA+ without the ontology, with partial ontology and with the complete ontology. In the first experiment, we implement the framework of the GCMP\_DF\_SVF based event classification scheme in Fig. 7. for 11 class events using the complete video (including pre-event, event-occ and post event). In this experiment, ontology is not utilized. The experimental results are show in the second column (GCMP+CNN+LSTM) in Table \RNum{10}. In the second experiment, we implement the framework of the semantic event classification scheme in Fig. 7. for 6-class events using the video segments including pre-event and event-occ. And we implement the frame work of success/failure classification algorithm in Fig. 8. using post-event video segments. In this experiment, the ontology is partly utilized. The final results are obtained by ensemble the results of 6-class event and success/failure classification. The results are shown in the third column (GCMP+CNN+LSTM and CNN) of Table \RNum{10}. In the third experiment, the proposed ontology based event classification scheme in Fig. 2. is implemented. In this experiment, the ontology is completely utilized. And the results are shown in the fourth column (On\_GCMP) of Table \RNum{10}. The experimental results show that the accuracy of experiment 3 is increased by 9.32\% than that of experiment 2 on average, and the experiment 2 is better by 10.74\% than experiment 1. It demonstrates that the scheme considering the ontology can improve the performance of the event classification.


\subsection{Comparison to state-of-the-art algorithm}
By combining the results of the two-stage event classification scheme in Section \RNum{6} D and the results of the success/failure classification algorithm in Section \RNum{6} E, we obtain the final results for all 11 events. In this section, we compare our algorithm with Ramanathan's work \cite{6} on NCAA, as shown in Table \RNum{11}. Ramanathan's results are obtained from Ref. \cite{6}. On-GCMP is our proposed scheme.

The experimental results demonstrate that our On\_GCMP algorithm on NCAA+ is more effective than Ramanathan's method on NCAA with evaluation metric of mean average precision. For single class events, the results of our algorithm is much better than Ramanathan's work for the three-point succ, free-throw fail, layup fail, other two-point succ, other two-point fail and slam dunk fail. The performance increases by 8.33\%--25.93\% for these events. Ramanathan's method obtained better results for three-point fail, layup success and slam dunk success and steal which are better than our scheme by 0.87\%--28.11\%. The proposed scheme and Ramanathan's work both get low performance on slam dunk event. It is possibly because the training set is imbalanced. The MAP of our scheme on NCAA+ is 58.10\%. It is higher by 6.50\% than Ramanathan's work on NCAA[11]. These results also show that our proposed On\_GCMP framework is effective for semantic event classification in basketball videos.

\begin{table}
\newcommand{\tabincell}[2]{\begin{tabular}{@{}#1@{}}#2\end{tabular}}
\centering
\footnotesize
\caption{Performance of success/failure classification algorithm on testing set I and II.}
\begin{tabular}{|c|c|c|c|} \hline
--&\--&\
Succ.&\
Fail.\\ \hline
\multirow{2}*{\tabincell{c}{Testing set II}}
&Number of event&148&134 \\
\cline{2-4}
&Accuracy(\%)&94.57&92.06 \\ \hline
\multirow{2}*{\tabincell{c}{Testing set I}}
&Number of event&757&1077 \\
\cline{2-4}
&Accuracy(\%)&99.6&99.26 \\ \hline
\end{tabular}
\end{table}

\subsection{Generalization ability of the proposed scheme}

In order to evaluate the generalization ability of the proposed scheme, we test the proposed scheme in testing set II. Testing set II is composed of videos from National Basketball Association (NBA) and Chinese Basketball Association (CBA) as shown in Section VI A, which are different from NCAA.

We first test the performance of the two-stage event classification scheme using pre-event and event-occ segments. The experimental results are shown in Table \RNum{12}.

From Table \RNum{12}, we can see that the trained model could get the similar accuracy in both testing set I and testing set II except for free throw and slam dunk events. For the slam dunk events, both testing set \RNum{1} and \RNum{2} get the poor performance. It is because the imbalance of training samples. Furthermore, we compare the free-throw in testing set II as shown in Fig. 9.

\begin{figure*}
\centering
\subfigure[From NCAA in testing set \RNum{1}]{
\label{Fig.sub.1}
\includegraphics[height = 0.5357in,width = 6in]{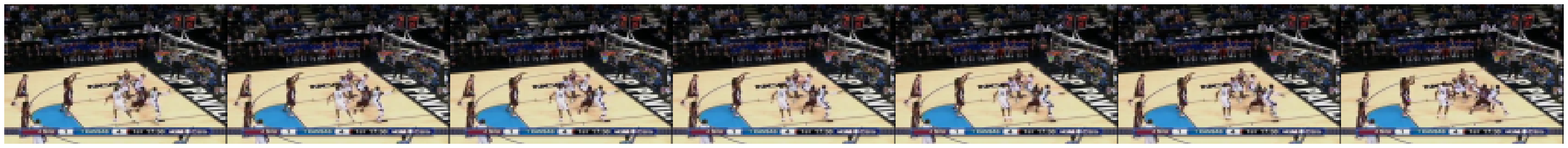}}
\subfigure[From CBA in testing set \RNum{2}]{
\label{Fig.sub.2}
\includegraphics[height = 0.5357in,width = 6in]{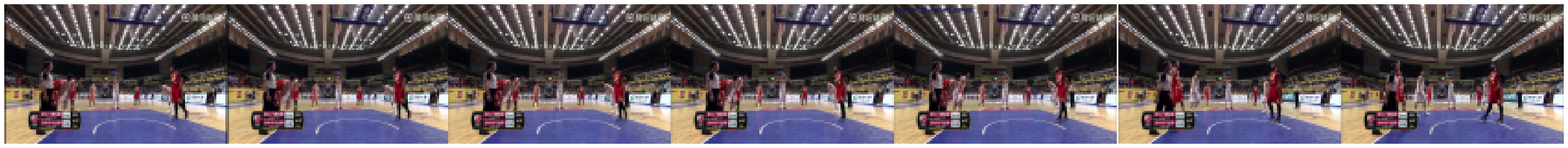}}
\subfigure[From NBA in testing set \RNum{2}]{
\label{Fig.sub.3}
\includegraphics[height = 0.5357in,width = 6in]{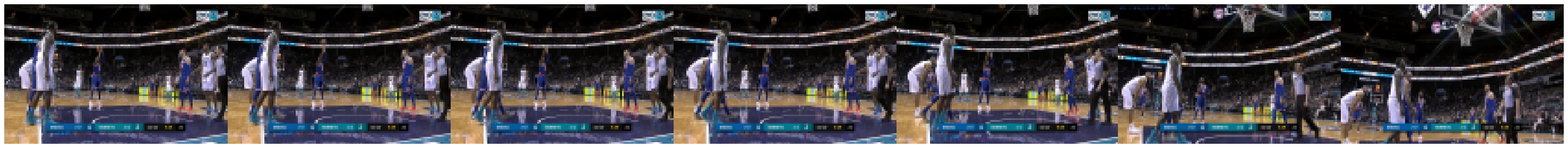}}
\caption{Example videos of free throw and slam dunk in in testing set \RNum{1} and \RNum{2}.}
\label{Fig.lable}
\end{figure*}

From Fig.9. we can see that that the presentation pattern for free-throw in testing set I and II are much different.

We further test the performance of the success/failure classification algorithms on testing set I and II respectively. The compared results are shown in Table \RNum{13}. From Table \RNum{13}, we can see that the proposed success/failure classification algorithm and the trained model are effective for both testing set I and testing set II, although the accuracy on testing set II is a little lower.

The experimental results in this section demonstrate that the observation and the corresponding proposed schemes are suitable to not only the NCAA game videos but also the NBA and CBA game videos. However, for some events such as free throw, the specific model should be trained. Furthermore, if the model could be fine tuned using the specific data, the accuracy could be further improved.

\subsection{Computational cost of our algorithm}
In this section, we evaluate the running time of our algorithm, as presented in Table \RNum{14}. The hardware was a computing server with one Nvidia TiTan X GPU. The main modules in the proposed scheme are event classification tasks on pre-event and event-occ segments and the success/failure prediction task on post-event segments. We report the running time of our scheme on these modules. Because there are various optic flow algorithms with different running times, the time cost does not include the running time of the optical flow algorithms.

Table \RNum{14} shows that the proposed method requires about 1.138 s to deal with 60-frame videos. In detail, the event classification using event-occ responsible for the five-category classification takes 0.485 s on average, the pre-event phase for layup/other two-point classification costs 0.351 s, and the final post-event phase for success/failure takes 0.302 s.

\section{Conclusion}
In this paper we propose an On-GCMP based scheme for event classification in basketball videos. By analyzing the correlation between the video clips of different events, we observe that the video clips in the NCAA dataset should be extended. Hence, we provide a novel dataset NCAA+, which is obtained automatically from NCAA. Using NCAA+, the proposed On\_GCMP scheme is tested for event classification. The experimental results show that the proposed scheme on NCAA+ can improve MAP by 6.5\% with respect to Ramanathan's work[11] on NCAA.

Although the proposed algorithm obtains the state-of-the-art performance on NCAA+ dataset, it is only 58.10\% which is far from a realistic application. In future, we try to introduce more information such as the pure collective motion pattern, individual poses, the players' locations and so on, so that the performance could be further improved for practical application. Furthermore, the slam dunk events get poor accuracy which is possibly because the training data is imbalanced. The training samples of slam dunk is too few to be represented effectively. It is an unsettled issue in this paper. In the future, we try to resolve it by revising the loss function with consideration of the data imbalance or trying other possible solutions.

\begin{table}
\centering
\footnotesize
\caption{Time cost for different phases of our method.}
\begin{tabular}{|c|c|c|c|c|} \hline
Model&Event-occ&Pre-event&Post-event&Total \\ \hline
Time cost(s) &0.485&0.351&0.302&1.138 \\ \hline
\end{tabular}
\end{table}

\ifCLASSOPTIONcaptionsoff
  \newpage
\fi


\begin{thebibliography}{1}
\bibitem{1} A. Krizhevsky, I. Sutskever, and G. Hinton, "Imagenet classification with deep convolutional neural networks",  \emph{International Conference on Neural Information Processing Systems}, vol. 60, pp. 1097-1106, 2012.

\bibitem{2} K. Simonyan and A. Zisserman, "Very deep convolutional networks for large-scale image recognition", \emph{International Conference on Learning Representations}, pp. 1-14, 2015.

\bibitem{3} C. Szegedy, W. Liu, Y. Jia, P. Sermanet, S. Reed, D. Anguelov, D. Erhan, V. Vanhoucke, and A. Rabinovich,  " Going deeper with convolutions," \emph{Proc. IEEE Conf. Computer Vision and Pattern Recognition}, pp. 1-9, 2015.

\bibitem{12} S. Ji, W. Xu, M. Yang, and K. Yu, "3d convolutional neural networks for human action recognition", IEEE Transactions on Pattern Analysis and Machine Intelligence, 2013, 35(1), pp. 221.


\bibitem{28} T. Brox, A. Bruhn, N. Papenberg, and J. Weickert, "Long-term Recurrent Convolutional Networks for Visual Recognition and Description", \emph{IEEE Trans. Pattern Analysis and Machine Intelligence}, vol. 39, no. 4, pp. 677, 2014.

\bibitem{29} D. Tran, L. Bourdev, R. Fergus, L. Torresani, and M. Paluri, "C3d: generic features for video analysis", arXiv preprint arXiv:1412.0767, 2014. 2, 5, 6.

\bibitem{30}  Y. H. Ng, M. Hausknecht, S. Vijayanarasimhan, O. Vinyals, R. Monga, and G. Toderici, "Beyond short snippets: Deep networks for video classification", \emph{Proc. IEEE Conf. Computer Vision and Pattern Recognition}, pp. 4694-4702, 2015.

\bibitem{31}  L. Yao, A. Torabi, K. Cho, N. Ballas, C. Pal, H. Larochelle, and A. Courville, "Describing videos by exploiting temporal structure", \emph{Proc. IEEE Conf. International Conference on Computer Vision}, pp. 4507-4515, 2015.

\bibitem{4} K. Soomro, A. RoshanZamir, and M. Shah, "UCF101: A dataset of 101 human actions classes from videos in the wild", CRCV-TR-12-01, 2012.1, 2.

\bibitem{5} G. Award, J. Fiscus, and W. Kraaij, "An overview of the goals, tasks, data, evaluation mechanisms and metrics", \hskip 1em plus 0.5em minus 0.4em\relax In TRECVID, 2014.1, 2.

\bibitem{6} V. Ramanathan, J. Huang, S. Abu-El-Haija, A. Gorban, K. Murphy, and F. Li, "Detecting events and key actors in multi-person videos", \emph{Proc. IEEE Conf. Computer Vision and Pattern Recognition}, pp.
    3043¨C3053, 2016.

\bibitem{32}  A. Gorban, H. Idrees, Y.G. Jiang, A. Roshan Zamir, I. Laptev, M. Shah, R. Sukthankar, "THUMOS challenge: Action recognition with a large number of classes", http://www.thumos.info/, 2015.

\bibitem{45} P. K. Rana, J. Taghia, Z. Ma, and M. Flierl, ¡°Probabilistic Multiview Depth Image Enhancement Using Variational Inference¡±, \emph{IEEE Journal of Selected Topics in Signal Processing (J-STSP)},Vol.9, No.3, pp.435¨C448, April 2015.

\bibitem{34}  Y.G. Jiang, Q. Dai, X. Xue, W. Liu, C.W. Ngo, "Trajectory-based modeling of human actions with motion reference points", \emph{European Conference on Computer Vision}, pp. 425-438, 2012.

\bibitem{35}  H. Wang, A. Klaser, C. Schmid, C. Lin, "Action Recognition by Dense Trajectories", \emph{Proc. IEEE Conf. Computer Vision and Pattern Recognition}, pp. 3169-3176, 2011.

\bibitem{36}  P. Wang, Y. Cao, C. Shen, L. Liu, H. T. Shen, "Temporal pyramid pooling based convolutional neural networks for action recognition", \emph{Proc. IEEE Conf. Computer Vision and Pattern Recognition}, pp. 2613-2622, 2017.

\bibitem{7}  M. S. Ibrahim, S. Muralidharan, Z. Deng, A. Vahdat, and G. Mori, "A hierarchical deep temporal model for group activity recognition", \emph{Proc. IEEE Conf. Computer Vision and Pattern Recognition}, pp. 1971-1980, 2016.

\bibitem{8} A. Karpathy, G. Toderici, S. Shetty, T. Leung, R. Sukthankar, and L. Fei-Fei, "Large-scale video classification with convolutional neural networks", \emph{Proc. IEEE Conf. Computer Vision and Pattern Recognition}, pp. 1725-1732, 2014.

\bibitem{9} L. Wang, Y. Xiong, Z. Wang, Y. Qiao, D. Lin, X. Tang, and L. V. Gool,  "Temporal segment networks: Towards good practices for deep action recognition", Acm Transactions on Information Systems, 22(1), pp. 20-36.

\bibitem{20} L. Kratz, and K. Nishino, "Anomaly detection in extremely crowded scenes using spatio-temporal motion pattern models", \emph{Proc. IEEE Conf. Computer Vision and Pattern Recognition}, pp. 1446-1453, 2009.

\bibitem{21} V. Mahadevan, W. Li, V. Bhalodia, and N. Vasconcelos, "Anomaly detection in crowded scenes", \emph{Proc. IEEE Conf. Computer Vision and Pattern Recognition}, pp. 1975-1981, 2010.

\bibitem{22} C. Loy, T. Xiang, and S. Gong, "Multi-camera activity correlation analysis", \emph{Proc. IEEE Conf. Computer Vision and Pattern Recognition}, pp. 1988¨C1995, 2009.

\bibitem{23} F. Xiong, X. Shi, D. Y. Yeung, "Spatiotemporal modeling for crowd counting in videos", arXiv preprint, arXiv:1707.07890, 2017.

\bibitem{38}  Y. H. Chen, and L. Y. Deng "Event Mining and Indexing in Basketball Video", \emph{Proc. IEEE Conf. International Conference on Genetic and Evolutionary Computing}, pp. 247-251, 2011.

\bibitem{39}  T. S. Fu, H. T. Chen, C. L. Chou, and W. J. Tsai, "Screen-strategy analysis in broadcast basketball video using player tracking", \emph{Proc. IEEE Conf. Visual Communications and Image Processing}, pp. 1-4, 2011.

\bibitem{37}  I. Atmosukarto, B. Ghanem, S. Ahuja, K. Muthuswamy, and N. Ahuja, "Automatic Recognition of Offensive Team Formation in American Football Plays", \emph{Proc. IEEE Conf. Computer Vision and Pattern Recognition}, pp. 991-998, 2013.

\bibitem{10} N. Dalal and B. Triggs, "Histograms of oriented gradients for human detection",\emph{Proc. IEEE Conf. Computer Vision and Pattern Recognition}, pp. 886-893, 2005.

\bibitem{11} I. Laptev, M. Marszalek, C. Schmid, and B. Rozenfeld. "Learning realistic human actions from movies", \emph{Proc. IEEE Conf. Computer Vision and Pattern Recognition}, pp. 1-8, 2008.

\bibitem{13} C. Feichtenhofer, A. Pinz, and A. Zisserman. "Convolutional two-stream network fusion for video action recognition", \emph{Proc. IEEE Conf. Computer Vision and Pattern Recognition}, pp. 1933-1941, 2016.

\bibitem{14} K. Simonyan and A. Zisserman. "Two-stream convolutional networks for action recognition in videos", \emph{Advances in Neural Information Processing Systems}, pp. 568-576, 2014.

\bibitem{15} Y. Wang, M. Long, J. Wang, and P. S. Yu, "Spatiotemporal Pyramid Network for Video Action Recognition", \emph{Proc. IEEE Conf. Computer Vision and Pattern Recognition}, pp. 2097-2106, 2017.


\bibitem{16} N. Vaswani, A. R. Chowdhury, and R. Chellappa, "Activity recognition using the dynamics of the configuration of interacting objects", \emph{Proc. IEEE Conf. Computer Vision and Pattern Recognition}, II-633-40 vol.2, 2003.

\bibitem{17} S. S. Intile, and A. F. Bobick, "Recognizing planned, multiperson action", \emph{Computer Vision and Image Understanding}, vol. 81, no. 3, pp. 414¨C445, 2001.

\bibitem{43} C. Pavel, H. Milan, V. Jan, and P. Josef, "Sports video classification in continuous TV broadcasts", \emph{Proc. IEEE Conf. International Conference on Signal Processing}, pp. 648-652, 2014.

\bibitem{18} D. Moore, and I. Essa, "Recognizing multitasked activities from video using stochastic context-free grammar", Proc.aaai National Conf. Proceedings of the National Conference on Artificial Intelligence, pp. 770-776, 2002.

\bibitem{19} S. M. Khan, and M. Shah, "Detecting group activities using rigidity of formation", ACM International Conference on Multimedia, pp. 403-406, 2005.

\bibitem{24} D.B. sam, S. Surya, and R.V. Babu, "Switching Convolutional Neural Network for Crowd Counting", \emph{Proc. IEEE Conf. Computer Vision and Pattern Recognition}, pp. 4031-4039, 2017.

\bibitem{25} T. Brox, A. Bruhn, N. Papenberg, and J. Weickert, "High accuracy optical flow estimation based on a theory for warping", \emph{European Conference on Computer Vision}, pp. 25-36, 2004.

\bibitem{44} Z. Ma, J. Xue, A. Leijon, Z. Tan, Z. Yang, and J. Guo, "Decorrelation of Neutral Vector Variables: Theory and Applications", \emph{IEEE Transactions on Neural Networks and Learning Systems}, Vol. 29, Issue 1, pp. 129-143, Jan. 2018.

\bibitem{27} S. Hochreiter, and J. Schmidhuber, "Long short-term memory", Neural computation, 9(8), 1997.

\bibitem{40}  L. Wu, J. He, M. Jian, S. Liu, and Y. Xu, "Global Motion Pattern based Event Recognition in Multi-person Videos", \emph{Proc. CCF Conf. Chinese Conference on Computer Vision}, pp. 667-676, 2017.

\bibitem{41} Y. Jia, E. Shelhamer, J. Donahue, S. Karayev, J. Long, R. B. Girshick, S. Guadarrama, and T. Darrell. "Caffe: Convolutional architecture for fast feature embedding", \emph{Proceedings of the ACM International Conference on Multimedia}, pp. 675¨C678, 2014.

\bibitem{42} R. Girdhar, D. Ramanan, A. Gupta, J. Sivic, and B. Russell, "ActionVLAD: Learning Spatio-Temporal Aggregation for Action Classification", \emph{Proc. IEEE Conf. Computer Vision and Pattern Recognition}, pp. 3165-3174, 2017.




\end{thebibliography}
\end{document}